\documentclass{article}

 \usepackage[main, final]{neurips_2026} 
 
\usepackage{graphicx}
\usepackage{amsmath}
\usepackage{array}
\usepackage{subcaption}
\usepackage[utf8]{inputenc} 
\usepackage[T1]{fontenc}    
\usepackage{hyperref}       
\usepackage{xcolor}  
\usepackage{url}            
\usepackage{booktabs}       
\usepackage{amsfonts}       
\usepackage{nicefrac}       
\usepackage{microtype}      
\usepackage{multirow}

\definecolor{mydarkblue}{RGB}{0,70,140}
\hypersetup{colorlinks=true,linkcolor=mydarkblue,citecolor=mydarkblue,urlcolor=mydarkblue}

\title{Log-Likelihood, Simpson's Paradox, and the Detection of Machine-Generated Text}

\author{
  Tom Kempton\thanks{Work completed while a consultant at the Risk and Security AI Lab, Visa Inc.} \\
  Department of Mathematics \\
  University of Manchester\\
  \texttt{tmwkempton@gmail.com} \\
  \And
  Viktor Drobnyi\\
  Risk and Security AI Lab \\
  Visa Inc. \\
  \texttt{vdrobnyi@visa.com} \\
  \AND 
  Maeve Madigan \\
  Risk and Security AI Lab \\
  Visa Inc. \\
  \texttt{mmadigan@visa.com} \\
  \And
  Stuart Burrell \\
  Risk and Security AI Lab \\
  Visa Inc. \\
  \texttt{sburrell@visa.com} \\
}

\begin{document}

\maketitle

\begin{abstract}
The ability to reliably distinguish human-written text from that generated by large language models is of profound societal importance. The dominant approach to this problem exploits the likelihood hypothesis: that machine-generated text should appear more probable to a detector language model than human-written text. However, we demonstrate that the token-level signal distinguishing human and machine text is non-uniform across the hidden space of the detector model, and naively averaging likelihood-based token scores across regions with fundamentally different statistical structure, as most detectors do, causes a form of Simpson's paradox: a strong local signal is destroyed by inappropriate aggregation. To correct for this, we introduce a learned local calibration step grounded in Bayesian decision theory. Rather than aggregating raw token scores, we first learn lightweight predictors of the score distributions conditioned on position in hidden space, and aggregate calibrated log-likelihood ratios instead. This single intervention dramatically and consistently improves detection performance across all baseline detectors and all datasets we consider. For example, our calibrated variant of Fast-DetectGPT improves AUROC from $0.63$ to $0.85$ on GPT-5.4 text, and a locally-calibrated DMAP detector we introduce achieves state-of-the-art performance. That said, our central contribution is not a new detector, but a precise diagnosis of a significant cause of underperformance of existing detectors and a principled, modular remedy compatible with any token-averaging pipeline. This will serve as a foundation for the community to build upon, with natural avenues including richer distributional models, improved calibration strategies, and principled ensembling with hidden-space geometry signals via the full Bayes-optimal decision rule.

\end{abstract}

\begin{figure}[ht]
    \centering
    \includegraphics[width=\linewidth]{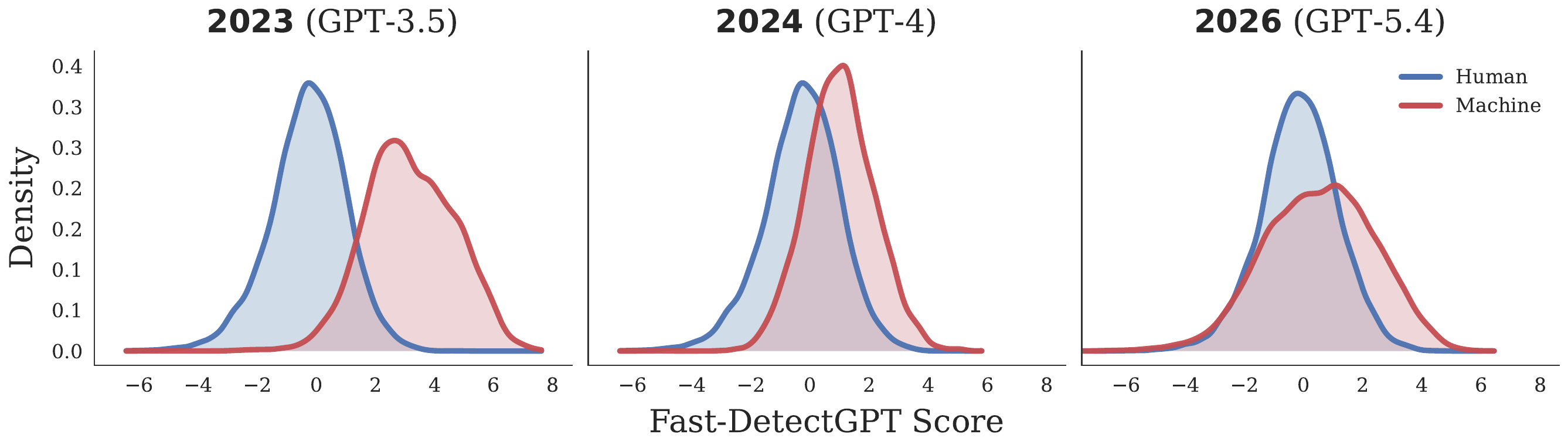}
    \caption{Performance of machine-generated text detectors appears to be degrading over time. These illustrative plots show Fast-DetectGPT \cite{baofast} scores for human-written and machine-generated text across three generations of instruction-tuned models: GPT-3.5, GPT-4, and GPT-5.4. The separation between human and machine score distributions deteriorates markedly across generations, with AUROC falling from \textbf{0.96} on GPT-3.5 to \textbf{0.63} on GPT-5.4. We attribute this to the progressive mitigation of overconfidence in successive frontier models, which weakens the token-level signal that likelihood-based detectors rely upon. Extensive experiments in Section~\ref{sec:experiments} and Appendix~\ref{app:further_results} provide further evidence of this degradation and highlight the need for effective new detection methodologies.}\label{fig:weakening_likelihood_hypothesis}
\end{figure}

\section{Introduction}\label{sec:intro}

The dominant approach to detecting machine-generated text rests on the \emph{likelihood hypothesis}: text produced by a language model should appear more probable under a detector model than human-written text does, reflecting the tendency of generators to favour high-probability tokens. This signal has driven a generation of detectors \citep{gehrmann2019gltr, mitchell2023detectgpt, baofast, hans2024spotting, su2023detectllm, zeng2024dald}, and considerable effort has gone into extracting it as effectively as possible.

In practice, however, the likelihood hypothesis is weakening, with performance of likelihood-based detectors severely degrading as the models used to generate text improve; see Figure~\ref{fig:weakening_likelihood_hypothesis} and Appendix~\ref{app:decay}. It was shown in \cite{kemptondmap} that the likelihood hypothesis fails as a method of detecting text generated by pure sampling from pretrained language models; when the likelihood hypothesis holds it is either because text has been generated using a sampling method such as top-$k$ or temperature explicitly designed to avoid low-probability tokens, or because text has been generated by an overconfident instruction-tuned language model. The question of optimally detecting top-$k$ and temperature-sampled text has been addressed in \cite{kempton2025temptest}; in this article we therefore focus on detecting text from instruction-tuned models, including modern closed-source frontier models such as GPT-5.4 \citep{openai2026gpt54}, Gemini 3.1 Pro \citep{google2026gemini31pro}, and Claude Sonnet 4.6 \citep{anthropic2026claudesonnet46}. In particular, our contribution is motivated by the following three recent advances in the literature, which we elaborate on further in Appendix~\ref{sec:ThreePoints}.
\begin{enumerate}
    \item {\bf Instruction-tuned models satisfy the likelihood hypothesis because of overconfidence.} It was argued in \cite{kemptondmap} that the reason the likelihood hypothesis continues to hold for instruction-tuned models is that the widely observed phenomenon of overconfidence caused by instruction tuning manifests as a token-level bias of instruction-tuned models towards tokens which a detector model finds likely.
    \item {\bf The degree of overconfidence varies across hidden space.} In \cite{xie2024calibrating} it was shown that overconfidence of instruction-tuned models does not present itself uniformly. In particular, it was shown that recalibrating models with temperature sampling is more effective when the chosen temperature varies as a function of the activation vector at the final hidden layer.
    \item {\bf Human and machine-generated text occupy different parts of hidden space.} In \cite{chen2025repreguard} it was shown that one can distinguish human and machine text by examining activation vectors in hidden space of a detector model, with human and machine-generated text typically occupying different regions of hidden space.
\end{enumerate}

\begin{figure}[!t]
\centering
\includegraphics[width=\linewidth]{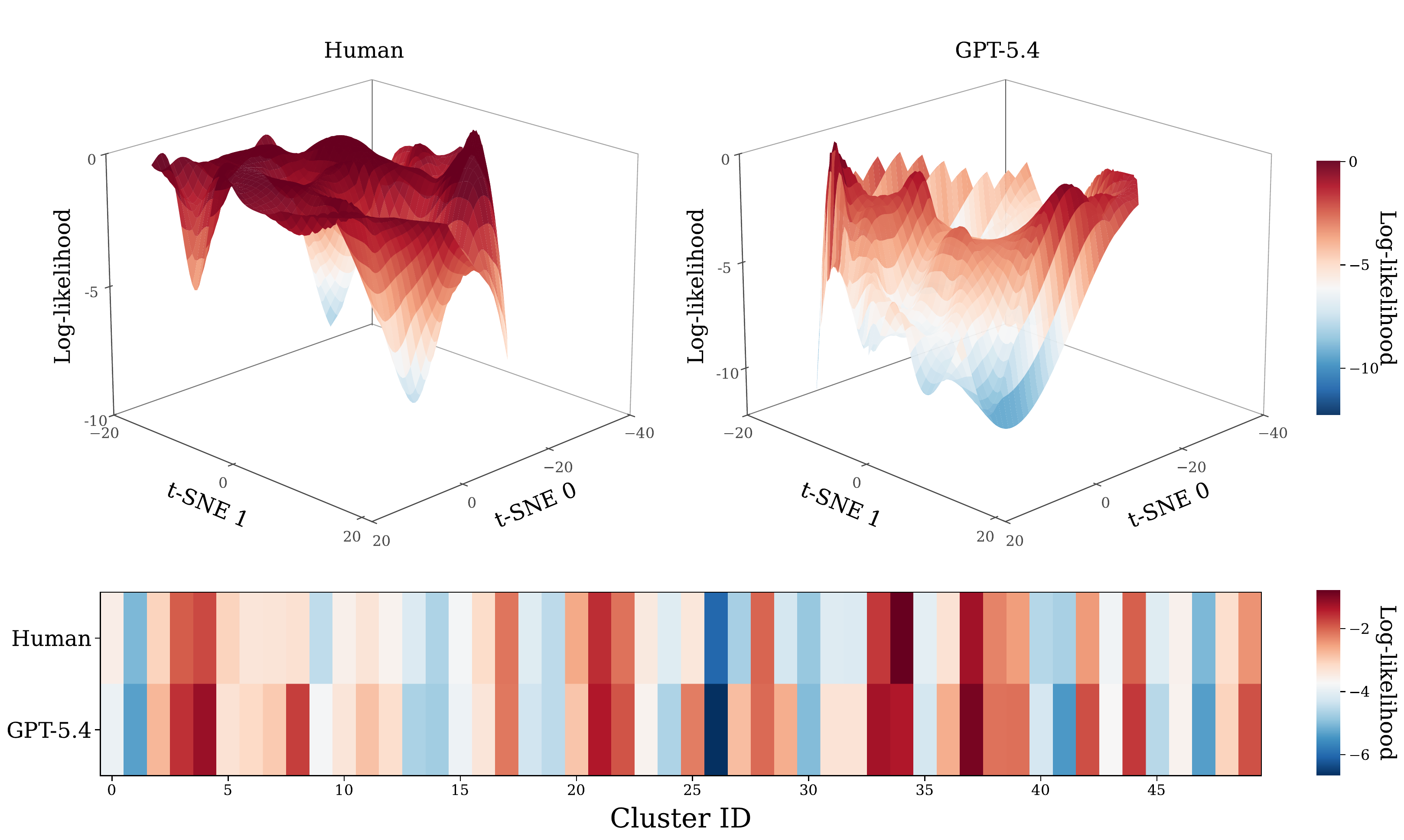}
\caption{\textbf{Upper:} A single illustrative example showing how the geometry of the log-likelihood landscape can vary across hidden space and source (human vs.\ GPT-5.4). \textbf{Lower:} Systematic analysis of 2,000 human texts and 2,000 GPT-5.4-generated texts from Modern RAID (a version of \cite{dugan2024raid} with newer models; see Section~\ref{sec:experiments}). While human text has lower average log-likelihood than GPT-5.4 ($-3.22$ vs.
$-3.05$), this relationship reverses in some regions of hidden space, and variation \emph{across} 50 $k$-means clusters exceeds variation \emph{between} sources. This suggests that contextualising token log-likelihood by hidden-space region provides a valuable untapped signal for detection. Further details on how to reproduce these plots are in Appendix~\ref{app:experimental_details}.}
\label{fig:cluster_heatmap}
\end{figure}

Together, these points suggest that we are at risk of a Simpson's paradox, where aggregation across heterogeneous subgroups destroys local signal. A classic illustration of this exists in healthcare: Hospital A may have a higher survival rate than Hospital B for both mild and severe cases of a disease, yet Hospital B could have a higher overall survival rate, simply because it treats proportionally more mild cases with a lower base rate of mortality. We show in Figure~\ref{fig:cluster_heatmap} that a similar phenomenon afflicts likelihood-based detectors: the token-level signal distinguishing human from machine text varies significantly across regions of hidden space, and naive averaging across those regions suppresses it because there are many regions dominating the aggregation where the signal is weaker or even reversed. Strong local signals exist, but current detectors fail to utilise them. Throughout this article we use the familiar term `Simpson's paradox' as shorthand for both true Simpson's paradoxes and the broader case of failure of aggregation under heterogeneous conditional score distributions.

\begin{figure}[!t]
\centering
\includegraphics[width=0.9\linewidth]{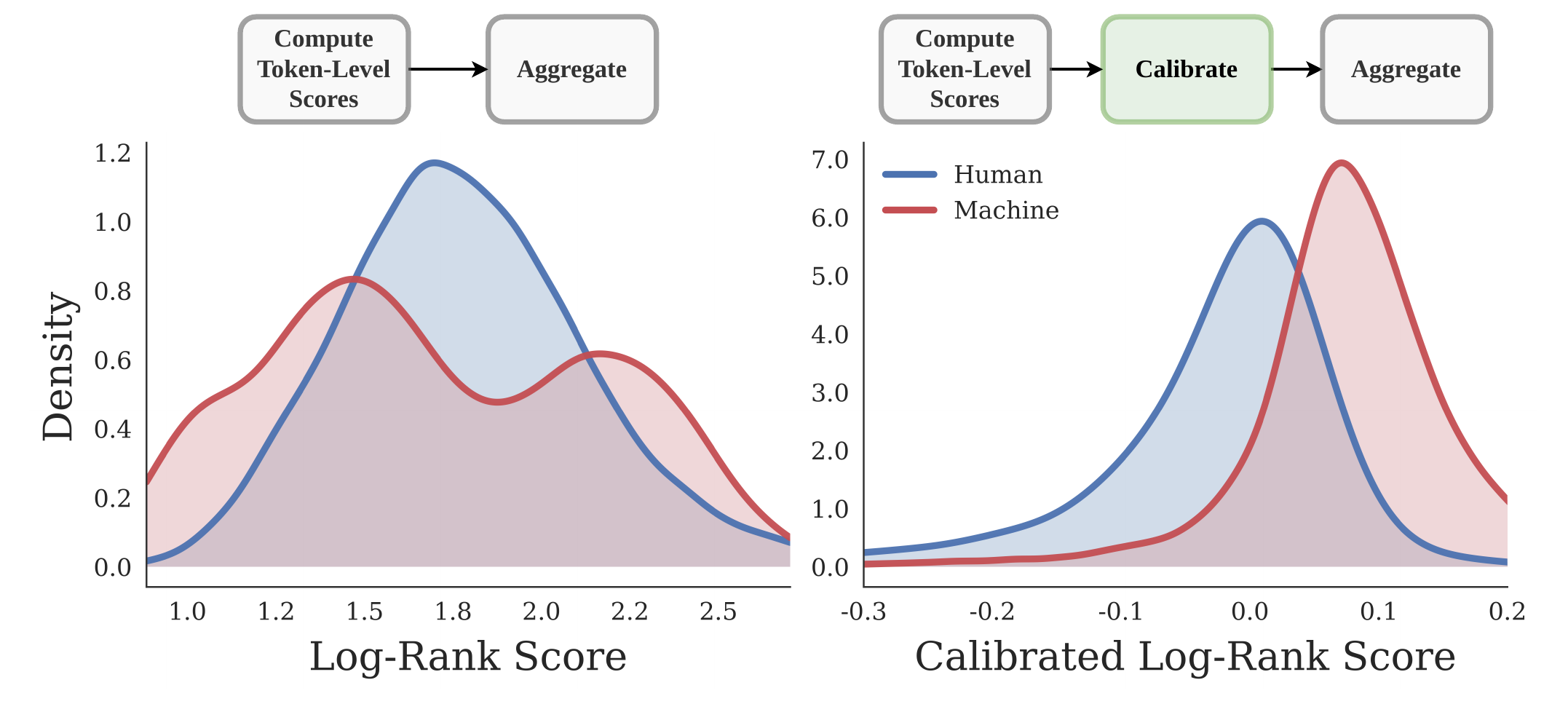}
\caption{\textbf{Left:} Standard detection pipelines aggregate raw per-token scores directly, which may yield heavily overlapping distributions for human and modern GPT-5.4 text (AUROC $0.56$). \textbf{Right:} Our pipeline inserts a local calibration step before aggregation, producing well-separated distributions (AUROC $0.91$). The improvement reflects a Simpson's paradox: token scores have heterogeneous local statistics across hidden-space regions, and naive aggregation obscures the underlying signal.}
\label{fig:workflow}
\end{figure}

We validate this idea by inserting an extra step into the standard pipeline for detecting machine-generated text. Instead of computing token-level scores, aggregating, and then thresholding, we add a learned local-calibration layer before aggregation, which prevents the problem of distortion due to different underlying statistics in different parts of hidden space; see Figure~\ref{fig:workflow}.

Our results strongly confirm that state-of-the-art detectors underperform due to the issues we describe, and that a learned local-calibration step dramatically improves performance across an array of state-of-the-art (SOTA) detectors. For example, we improve the AUROC of Fast-DetectGPT \cite{baofast} from 0.63 to 0.85 on GPT-5.4-generated texts by incorporating the local-calibration step. Our best-performing approach, a locally-calibrated detector based on DMAP \cite{kemptondmap}, outperforms all baselines, and we have open-sourced the implementation on GitHub.\footnote{\url{https://github.com/TMKempton/calibrated-text-detection}}

\paragraph{Contributions}
\begin{itemize}
    \item \textbf{Demonstrate the weakening of the likelihood hypothesis.} We demonstrate that the performance of modern detectors is diminishing as overconfidence is mitigated in frontier models. 
    \item \textbf{Identify a major cause of underperformance in modern detectors.} We demonstrate how a phenomenon resembling Simpson's paradox leads to underperformance in modern detectors due to non-uniformity of token-level scores across hidden space.
    \item \textbf{Introduce learnable local calibration as a mitigation strategy.} Grounded in Bayesian decision theory, we derive a principled calibration step that learns local estimates of token-score distributions conditioned on position in hidden space, and aggregates calibrated log-likelihood ratios in place of raw token scores.
    \item \textbf{Validate that learned local calibrators significantly improve detection.} Across all baseline detectors and datasets considered, inserting a local calibration step before score aggregation yields substantial and consistent gains in AUROC, with the locally-calibrated DMAP detector achieving state-of-the-art performance.
\end{itemize}

We emphasise that our primary goal is not to propose a new detector, but to isolate and correct a fundamental flaw in how existing detectors aggregate information. The local calibration step we introduce is deliberately lightweight, theoretically grounded, and architecture-agnostic: it can be inserted into any token-averaging detection pipeline with minimal overhead. We intentionally avoid extensive hyperparameter tuning or dataset-specific optimisation, precisely because we want to demonstrate that the gains arise from correcting a statistical pathology rather than from over-engineering to a particular benchmark. By identifying a root cause of detector underperformance and providing a modular remedy, we offer the community a foundation to build upon with further optimisation.

\section{Related Work}

\paragraph{Likelihood-based detection.} The most widely used signal in machine-generated text detectors is per-token log-likelihood \cite{gehrmann2019gltr}, which takes a text $w_1\cdots w_n$ and a detector language model $p$ and computes $\frac{1}{n}\sum_{i=1}^n \log(p(w_i|w_1\cdots w_{i-1}))$. Per-token log-rank is similar, replacing likelihood with the rank of token $w_i$ in the vocabulary ordered by decreasing probability under $p(\cdot|w_1\cdots w_{i-1})$. Both exploit the likelihood hypothesis: machine-generated text should sit higher in the probability distribution of a detector model than human-written text.

\paragraph{Contextualisation of token-level scores.} A natural limitation of raw likelihood and log-rank scores is that they vary substantially across writing styles and domains; the next token in a piece of poetry is inherently less predictable than in a chemistry textbook. DetectGPT \cite{mitchell2023detectgpt} addressed this by subtracting from the log-likelihood of a text the expected log-likelihood of alternative phrasings of the same content. Fast-DetectGPT, DetectLLM, Binoculars, and DMAP all build on this contextualisation idea \cite{baofast, su2023detectllm, hans2024spotting, kemptondmap}. While DMAP was not originally proposed as a detector of machine-generated text, it presents a contextualised generalisation of log-rank with well-understood statistical properties that makes it a natural fit for our framework; we adapt it to produce a detector in Appendix~\ref{sec:AdaptingDMAP}.

\paragraph{Complementary approaches.} Methods that exploit patterns in the log-likelihood sequence \cite{sun2026ai, luo2026specdetect}, statistical signatures of specific decoding strategies \cite{kempton2025temptest}, and geometric differences between human and machine text in embedding space \cite{tulchinskii2024intrinsic, chen2025repreguard} offer signals that are largely orthogonal to the likelihood signal, making them natural candidates for ensembling with likelihood-based methods.

\paragraph{Zero-shot versus supervised detection.} {The likelihood baselines above are zero-shot score functions: no labelled examples are required to produce a score, although a labelled set is still needed to choose an operating threshold. DALD \cite{zeng2024dald} is supervised: it fine-tunes the detector model on text from the target generator, then applies Fast-DetectGPT with the fine-tuned model. Our method is also supervised, since it learns generator- and domain-dependent conditional score distributions from labelled human and machine text. The distinction is where supervision enters: DALD changes the detector model, whereas our method calibrates how token-level scores are interpreted. Table~\ref{tab:calibration_supervision} isolates these effects, and Appendix~\ref{app:dald_comparison} discusses them alongside token-budget results.}

\section{Calibrating Score Functions with Bayesian Decision Theory}\label{sec:calibration_theory}

Suppose we have a text $w_1\cdots w_n$ which we wish to classify as human-written or machine-generated. Let $x=w_1\cdots w_n$ and $x_i=w_i|w_1\cdots w_{i-1}$.

\paragraph{The standard pipeline.} Token-level score functions work as follows. The text is run through a detector language model and each token is assigned a score $g(x_i)$, as described in Section~\ref{sec:baselines}. These scores are aggregated to produce a text-level score $\frac{1}{n}\sum_{i=1}^n g(x_i)$, which is then thresholded against a value learned from a labelled set of human-written and machine-generated texts.

\paragraph{Naive aggregation.} This aggregation of $g$ is optimal only if the distribution of token scores $g(x_i)$ is independent of any other information available to us. As argued in the introduction, this assumption fails: the statistical distribution of token-level scores varies significantly across regions of hidden space, and aggregating naively across these regions suppresses the local signal. This is precisely the phenomenon identified in Section~\ref{sec:intro} where inappropriate averaging of heterogeneous local score distributions weakens the global signal.

To see this more formally, suppose the only information available about a text is its aggregate score $\alpha = \frac{1}{n}\sum_{i=1}^n g(x_i)$. Let $H$ and $M$ denote the hypothesis that the text is human-written or machine-generated, respectively. The Neyman--Pearson lemma states that the optimal test thresholds the log-likelihood ratio
\[
\Lambda_1(\alpha) = \log\left(\frac{\mathbb{P}(\text{score} = \alpha \mid H)}{\mathbb{P}(\text{score} = \alpha \mid M)}\right).
\]
Given the reasonably smooth score distributions shown in Figure~\ref{fig:weakening_likelihood_hypothesis}, thresholding directly on $\alpha$ is a reasonable proxy for this. However, the aggregate score is not the only information available to us, and exploiting additional information can substantially improve detection.

\paragraph{The Bayes-optimal decision rule.} Let $Z(x_i)$ denote additional features extracted at the token level, specifically a low-dimensional projection of the activation vector at the final hidden layer of the detector model, together with properties of the next-token probability distribution $p(\cdot \mid w_1\cdots w_{i-1})$. The precise construction of $Z(x_i)$ is described in Section~\ref{sec:calibration_implementation}.

Given token-level scores $g(x_i)$ and features $Z(x_i)$, the Bayes-optimal decision rule is to threshold the log-likelihood ratio of the full observed data,
\[
\Lambda_2 = \log\left(\frac{\mathbb{P}(\{g(x_i), Z(x_i)\}_{i=1}^n \mid H)}{\mathbb{P}(\{g(x_i), Z(x_i)\}_{i=1}^n \mid M)}\right).
\]
Learning the full sequence distributions $\mathbb{P}(\cdot \mid H)$ and $\mathbb{P}(\cdot \mid M)$ is challenging. Assuming instead that token-level scores are conditionally independent of one another given the features, and that hidden-space features are independent across tokens, the Bayes-optimal decision rule decomposes as
\begin{align}
\Lambda_3 &= \sum_{i=1}^n \log\left(\frac{\mathbb{P}(g(x_i), Z(x_i) \mid H)}{\mathbb{P}(g(x_i), Z(x_i) \mid M)}\right) \notag \\
&= \sum_{i=1}^n \left(\log\left(\frac{\mathbb{P}(g(x_i) \mid Z(x_i), H)}{\mathbb{P}(g(x_i) \mid Z(x_i), M)}\right) + \log\left(\frac{\mathbb{P}(Z(x_i) \mid H)}{\mathbb{P}(Z(x_i) \mid M)}\right)\right). \label{eq:Bayes-Opt}
\end{align}

Note that these conditional independence assumptions are unlikely to hold precisely, and there is likely significant room for improvement by relaxing them; see Section~\ref{sec:tuning}.

\paragraph{Interpreting the decomposition.} The two terms in Equation~\eqref{eq:Bayes-Opt} capture complementary sources of signal. The first term is a locally-calibrated likelihood ratio: it asks how probable the token score $g(x_i)$ is, given the local context $Z(x_i)$ and the hypothesis that the text is human-written or machine-generated. This is the signal that naive aggregation suppresses. The second term captures how much the position of the token in hidden space itself discriminates between human and machine text, independently of the score; this is the principle underlying the detector of \cite{chen2025repreguard}.

\paragraph{Our calibrated score.} The purpose of this paper is to demonstrate the power of calibrating existing score functions using local information $Z(x_i)$, rather than to ensemble likelihood-based scores with hidden-space geometry signals. Our calibration step therefore uses only the first term of Equation~\eqref{eq:Bayes-Opt}. For a fixed generator model $M$, we learn plug-in estimates of $\mathbb{P}(g(x_i) \mid Z(x_i), H)$ and $\mathbb{P}(g(x_i) \mid Z(x_i), M)$ and compute their log-ratio, returning the calibrated score
\[
\Lambda_4 = \sum_{i=1}^n \log\left(\frac{\mathbb{P}(g(x_i) \mid Z(x_i), H)}{\mathbb{P}(g(x_i) \mid Z(x_i), M)}\right).
\]
The implementation of these plug-in estimators is described in Section~\ref{sec:calibration_implementation}.

\section{Implementing Local Calibration}\label{sec:calibration_implementation}

In this section we describe our implementation of the local calibration step, and in particular how we learn plug-in estimates for the probabilities $\mathbb{P}(g(x_i)|Z(x_i),H)$ and $\mathbb{P}(g(x_i)|Z(x_i),M)$. In deployment, careful thought should be given to the design of the calibration step, based on the specific score function whose performance one seeks to optimise, the nature of the texts one wishes to analyse, and the desired balance between false positives and false negatives. For example, one may wish to place upper and lower bounds on the contribution of any single token so that the aggregate score is not dominated by a handful of extreme values.

{Our goal is to test whether local calibration improves likelihood-based scores. We deliberately begin with a simple implementation: a local Gaussian predictor for each scalar score function, learned at different positions in hidden space with no detector-specific tuning. Section~\ref{sec:experiments} shows consistent in-distribution AUROC gains, but results at low false-positive rates (FPR) and under domain shift expose limitations of the Gaussian assumption and calibration coverage; Section~\ref{sec:tuning} discusses improvements.}

\subsection{Baseline Score Functions}\label{sec:baselines}

Our focus is on extracting signal based on the likelihood hypothesis. We use as baselines the most successful approaches in this family: Fast-DetectGPT \cite{baofast}, Binoculars \cite{hans2024spotting}, NPR \cite{su2023detectllm}, per-token log-surprisal, and per-token log-rank \cite{gehrmann2019gltr}. Approaches that use the likelihood signal in complementary ways, for example by examining local fluctuations or differences between the first and second halves of a text, are orthogonal to our investigation and are not included as baselines.

For our calibration step to apply, a detector must operate by averaging a token-level score function $g$ over the tokens in a text. Per-token log-likelihood sets \[g(w_i|w_1\cdots w_{i-1})=\log(p(w_i|w_1\cdots w_{i-1}))\] and computes $\frac{1}{n}\sum_{i=1}^n g(w_i|w_1\cdots w_{i-1})$. Per-token log-rank sets \[g(w_i|w_1\cdots w_{i-1})=\log(r(w_i|w_1\cdots w_{i-1})),\] where $r$ denotes the rank of the token under $p$.

NPR, Binoculars, and Fast-DetectGPT are not direct averages of token-level scores. For example, Fast-DetectGPT sums log-likelihood and entropy over a text but then normalises by the variance of the scores rather than the token count. For these baselines we use the true function as the reported baseline score, but train our locally-calibrated variant using a token-level approximation. Full details of each baseline and its token-level counterpart are given in Appendix~\ref{sec:scoring}.

\subsection{Learning Local Score Distributions}

Given a sequence of tokens $w_1\cdots w_{i-1}$ as context, we define the \textit{hidden vector} as the activation vector at the final hidden layer of the detector model when processing that context. We first perform PCA on the hidden vectors to identify $d$ principal directions in hidden space (in our experiments $d=25$). For each context in the training set, the feature vector $Z(x_i)$ is then formed by concatenating this $d$-dimensional projection of the hidden vector with the model probabilities of the top-$k$ (we set $k=5$ in our experiments) candidate tokens, yielding a $(d+k)$-dimensional feature vector. We then train, for each generator model (including human), a local DMAP predictor and local Gaussian predictors of other score functions using a simple two-layer MLP as described in Appendix~\ref{app:experimental_details}.

\paragraph{Computing the calibrated score.} Given a context, a true next token, and plug-in estimators for human and machine distributions, we compute $\Lambda_4$ by subtracting the log-loss of the human predictor from that of the machine predictor, token by token, and summing over the text. This yields the locally-calibrated score defined in Section~\ref{sec:calibration_theory}.

\section{Experiments}\label{sec:experiments}

\subsection{Experimental Setup}

We evaluate on three datasets spanning different domains, generators, and time periods. Classic RAID \cite{dugan2024raid} pairs human-written texts across eight domains with machine-generated counterparts from GPT-3.5 and GPT-4, generated via pure sampling without evasion attacks. Modern RAID extends this to current frontier models by regenerating responses following the original prompt templates using GPT-5.4 \cite{openai2026gpt54}, Gemini 3.1 Pro \cite{google2026gemini31pro}, and Claude Sonnet 4.6 \cite{anthropic2026claudesonnet46} with default sampling settings. Finally, the Peer-Review benchmark \cite{yu2025your} contains reviews of ICLR and NeurIPS submissions (2021--2022) generated by GPT-4o, Claude 3.5 Sonnet, and Gemini 1.5 Pro.

We use OPT-125M as the detector model in all experiments except for the Binoculars baseline, for which we used the Falcon-3-7B scorer--observer pair. For each dataset and generator, we hold out a fixed subset for training local predictors, with the remainder used for evaluation. We report AUROC as our primary metric; by the Mann--Whitney interpretation, AUROC is the probability that a randomly chosen machine-generated text receives a higher detector score than a randomly chosen human-written text. Our main results use only the first 200 tokens of each text to facilitate direct and fair comparison; additional experimental details are provided in Appendix~\ref{app:experimental_details}. {Appendix~\ref{app:token_budgets} reports AUROC and true-positive rate (TPR) at 0.1\% and 1\% FPR across token budgets from 50 to 400 tokens.}

\subsection{Discussion of Results}

\begin{table}[t]
\centering
\caption{Token-level scoring functions are substantially improved by adding a local calibration step. This contextualizes the score based on the location in hidden space. Confidence intervals for all values are provided by the companion Table \ref{tab:calibration_comparison_cis} in Appendix \ref{app:further_results}.}
\label{tab:calibration_comparison}
\vspace{0.5em}
\resizebox{\textwidth}{!}{
\begin{tabular}{lcccccc}
\toprule
\multirow{2}{*}{Method}
& \multicolumn{3}{c}{Modern RAID}
& \multicolumn{3}{c}{Peer-Review} \\
\cmidrule(lr){2-4} \cmidrule(lr){5-7}
& GPT-5.4 & Gemini & Claude
& GPT-4o & Gemini & Claude \\
\midrule

Baseline Log-Surprisal
& 0.562 
& 0.506 
& 0.631 
& 0.635 
& 0.559 
& 0.514 \\

Calibrated Log-Surprisal
& \textbf{0.918}
& \textbf{0.893} 
& \textbf{0.937 }
& \textbf{0.954} 
& \textbf{0.969} 
& \textbf{0.990} \\

\midrule

Baseline Log-Rank
& 0.544 
& 0.468 
& 0.608 
& 0.602 
& 0.527 
& 0.479 \\

Calibrated Log-Rank
& \textbf{0.856} 
& \textbf{0.781} 
& \textbf{0.869 }
& \textbf{0.826} 
& \textbf{0.835} 
& \textbf{0.910} \\

\midrule

Baseline Token NPR
& 0.603 
& 0.686 
& 0.768 
& 0.616 
& 0.584 
& 0.537 \\

Calibrated Token NPR
& \textbf{0.834} 
& \textbf{0.749}
& \textbf{0.856} 
& \textbf{0.809} 
& \textbf{0.813} 
& \textbf{0.900} \\

\midrule

Baseline Token Fast-DetectGPT
& 0.628 
& 0.734 
& 0.800 
& 0.640 
& 0.607 
& 0.545 \\

Calibrated Token Fast-DetectGPT
& \textbf{0.854} 
& \textbf{0.805} 
& \textbf{0.868} 
& \textbf{0.862} 
& \textbf{0.861} 
& \textbf{0.932} \\

\midrule 

Baseline DMAP
& 0.689 
& 0.779 
& 0.805 
& 0.834 
& 0.952 
& 0.861 \\

Calibrated DMAP
& \textbf{0.939} 
& \textbf{0.937} 
& \textbf{0.961} 
& \textbf{0.973} 
& \textbf{0.979} 
& \textbf{0.994} \\

\bottomrule
\end{tabular}}
\end{table}

\begin{table}[t]
\centering
\caption{{Comparison with zero-shot score functions and supervised methods. Calibrated DMAP has the highest AUROC in five of six generator-specific settings; the supervised DALD leads on Modern RAID Gemini. Confidence intervals are in Table \ref{tab:baseline_comparison_cis}.}}
\label{tab:baseline_comparison}
\vspace{0.5em}
\begin{tabular}{lcccccc}
\toprule
\multirow{2}{*}{Method}
& \multicolumn{3}{c}{Modern RAID}
& \multicolumn{3}{c}{Peer-Review} \\
\cmidrule(lr){2-4} \cmidrule(lr){5-7}
& GPT-5.4 & Gemini & Claude
& GPT-4o & Gemini & Claude \\
\midrule

NPR
& 0.526 
& 0.503 
& 0.554 
& 0.600 
& 0.582 
& 0.496 \\

Binoculars
& 0.692 
& 0.537 
& 0.628 
& 0.289 
& 0.626 
& 0.508 \\

Fast-DetectGPT
& 0.629 
& 0.733 
& 0.799 
& 0.640 
& 0.606 
& 0.543 \\

DALD
& 0.894 
& \textbf{0.957} 
& 0.943 
& 0.852 
& 0.865 
& 0.965 \\

Calibrated DMAP
& \textbf{0.939} 
& 0.937 
& \textbf{0.961} 
& \textbf{0.973} 
& \textbf{0.979}
& \textbf{0.994}  \\

\bottomrule
\end{tabular}
\end{table}

\begin{table}[t]
\centering
\caption{{Calibration-versus-supervision ablation on Modern RAID at 200 tokens. All DMAP variants use the optimised DMAP configuration of Appendix \ref{app:optimising_dmap}. DALD and the supervised DMAP variants use labelled generator-specific data; calibrated variants additionally condition score interpretation on hidden-space features. Best values within each generator are bold. These results confirm that the uplift due to our calibration method is not merely the effect of supervision.}}
\label{tab:calibration_supervision}
\vspace{0.5em}
{
{\begin{tabular}{llccc}
\toprule
Generator & Method & TPR@0.1\% & TPR@1\% & AUROC \\
\midrule
\multirow{5}{*}{GPT-5.4} & DMAP & 0.85 & 6.75 & 0.689 \\
& DALD & 30.2 & 42.7 & 0.894 \\
& Supervised DMAP & 5.21 & 29.4 & 0.904 \\
& Calibrated DMAP & 39.0 & 67.1 & 0.967 \\
& Calibrated Supervised DMAP & \textbf{41.4} & \textbf{71.7} & \textbf{0.978} \\
\midrule
\multirow{5}{*}{Gemini} & DMAP & 0.07 & 1.59 & 0.779 \\
& DALD & 34.7 & 56.4 & 0.957 \\
& Supervised DMAP & 0.92 & 7.87 & 0.873 \\
& Calibrated DMAP & 39.8 & 63.0 & 0.952 \\
& Calibrated Supervised DMAP & \textbf{41.7} & \textbf{70.5} & \textbf{0.971} \\
\midrule
\multirow{5}{*}{Claude} & DMAP & 0.84 & 7.35 & 0.805 \\
& DALD & 27.0 & 48.5 & 0.943 \\
& Supervised DMAP & 4.51 & 27.6 & 0.926 \\
& Calibrated DMAP & 50.5 & 73.0 & 0.975 \\
& Calibrated Supervised DMAP & \textbf{50.8} & \textbf{82.9} & \textbf{0.987} \\
\bottomrule
\end{tabular}}
}
\end{table}


{Table~\ref{tab:calibration_comparison} compares token-level scorers with calibrated counterparts, and Table~\ref{tab:baseline_comparison} compares calibrated DMAP with zero-shot and supervised baselines. Tables \ref{tab:classic_raid_key_metrics}, \ref{tab:modern_raid_key_metrics}, and \ref{tab:peer_review_key_metrics} report confidence intervals and low-FPR results. Appendix~\ref{app:token_budgets} additionally reports AUROC, and TPR at 0.1\% and 1\% FPR from 50 to 400 tokens.}

\paragraph{{Local calibration improves in-distribution AUROC.}} {Table~\ref{tab:calibration_comparison} shows an AUROC increase for every scorer, generator, and in-distribution dataset evaluated (30/30 comparisons). On GPT-5.4, calibrated log-surprisal rises from $0.562$ to $0.918$, and calibrated DMAP from $0.689$ to $0.939$; on the Peer-Review Claude split, calibrated Token Fast-DetectGPT rises from $0.545$ to $0.932$.}

\paragraph{{Performance at very low false-positive rates.}} {Tables \ref{tab:classic_raid_key_metrics}--\ref{tab:peer_review_key_metrics} show that AUROC improvements do not always transfer uniformly to extreme operating points. This is primarily a weakness of the Gaussian conditional model used by some calibrated scorers, rather than of calibration itself; Appendix~\ref{sec:gaussian_assumption} documents skew and asymmetric tails that are especially consequential at stringent thresholds. DMAP makes no Gaussian assumption in the calibration step, and performs well at low FPR.} 

\paragraph{Calibration versus supervision.} Our calibration step adds an element of supervision to detectors, and one may wonder to what extent our results really show the effectiveness of calibration as opposed to supervision more generally. We study this in Table~\ref{tab:calibration_supervision}, where we introduce supervised DMAP, an uncalibrated DMAP detector based on a version of OPT-125M fine-tuned on outputs of the generating model (following the recipe of DALD). We see that this supervision is significantly inferior to our calibration step and conclude that calibration offers very substantial benefit beyond generic supervision. Interestingly, applying this fine-tuning in addition to our calibration makes the detector even stronger. We also see in this table that our calibrated DMAP detector is stronger than DALD, including at low false-positive rates. This table uses the optimised calibration setup for DMAP described in Appendix~\ref{app:optimising_dmap}. An earlier version of this work reported results for optimised DMAP only in the appendix; they are presented in the main text here in response to referee concerns that calibrated DMAP may not outperform DALD in all settings. 

\paragraph{{Dependence on calibration coverage.}} {These results are supervised and depend on calibration data that represents the deployed generator, domain, and relevant population factors. Under the pooled-generator setting, DALD leads calibrated DMAP on Modern RAID (AUROC $0.924$ vs.\ $0.904$; Table~\ref{tab:intel_modern_multigenerator_partial}), and leave-one-domain-out evaluation produces typical 8--12 point AUROC drops (Table~\ref{tab:domain_shift_id_ood}). A broad calibration mixture can reduce this mismatch, but performance on an unseen generator or domain should not be inferred from the in-distribution tables. This caveat is important because current zero-shot likelihood baselines are not competitive on the frontier-model settings studied here, so practical systems may need supervision while explicitly monitoring its coverage.}

\section{Future Directions}\label{sec:tuning}

{Throughout this paper we adopt a simple experimental approach, restricting ourselves to likelihood-derived signals and limited tuning. Our goal is to test whether a basic calibration step recovers signal. We identify five directions for improving accuracy, robustness, and deployment reliability.}

\paragraph{Richer distributional models.} The most significant limitation of our current implementation is the use of simple Gaussians to model $\mathbb{P}(g(x_i)|Z(x_i),H)$ and $\mathbb{P}(g(x_i)|Z(x_i),M)$. These are clearly suboptimal; log-rank, for example, takes non-negative discrete values with the most common value being $0$, making Gaussian approximations a poor fit. The weakness of the Gaussian assumption for log-likelihood is examined in Appendix~\ref{sec:gaussian_assumption}. Replacing these with more expressive density estimators, such as normalising flows or mixture models, would be a natural first step.

\paragraph{Robustness to imperfect calibration.} Bayes-optimal decision theory assumes that the class-conditional distributions are known exactly. In practice, our plug-in estimators are imperfect, and the consequences of this imperfection depend on the relative cost of false positives and false negatives, the statistics of the dataset, and the properties of the score function. Principled interventions, such as capping the per-token contribution to the aggregate score to limit the influence of poorly estimated tails, could improve robustness and deserve investigation.

\paragraph{{Coverage-aware confidence.}} A deployed detector should report whether a test text lies within the support of its calibration data. One practical direction is a hidden-space coverage score based on distances or density under the training features $Z(x_i)$, aggregated over tokens and calibrated on held-out data. Low coverage would trigger abstention, broader calibration, or human review rather than an overconfident classification. Developing such a measure, and testing whether it predicts generator, domain, and demographic shift, is an important next step. Reporting conditional AUROC within hidden-space clusters would provide a complementary diagnostic of where local signal is gained or lost.

\paragraph{Incorporating hidden-space geometry.} In deriving $\Lambda_4$ we deliberately set aside the second term of Equation~\eqref{eq:Bayes-Opt}, which captures how much the position of a token in hidden space itself discriminates between human and machine text. Given the effectiveness of \cite{chen2025repreguard}, we expect that learning plug-in estimates of $\mathbb{P}(Z(x_i)|H)$ and $\mathbb{P}(Z(x_i)|M)$ and incorporating them into the decision rule would yield substantial further gains, effectively ensembling our calibrated likelihood signal with a hidden-space geometry signal in a principled, Bayes-optimal manner.

\paragraph{Relaxing conditional independence.} In moving from $\Lambda_2$ to $\Lambda_3$ we assume that token-level scores are conditionally independent given features. This assumption is known to fail: several detectors are explicitly built around the sequential dependencies in the log-likelihood sequence \cite{sun2026ai, luo2026specdetect}. With sufficient training data, learning the full conditional distributions in $\Lambda_2$ directly, rather than factorising over tokens, would be expected to yield further improvements.

\section{Conclusion}

The performance of likelihood-based detectors of machine-generated text is degrading rapidly as the overconfidence of instruction-tuned frontier models is reduced. For example, we see that Fast-DetectGPT, representative of the state of the art, achieves an AUROC of $0.96$ on GPT-3.5 text, $0.79$ on GPT-4 text, and $0.63$ on GPT-5.4 text. 

We have identified a significant and previously unrecognised cause: naive aggregation of token-level scores across regions of hidden space with fundamentally different statistical structure constitutes a form of Simpson's paradox, suppressing a local signal that remains strong. The fix is principled and modular. By inserting a learned local calibration step between token scoring and score aggregation, grounded in Bayesian decision theory, we recover dramatic performance improvements across every baseline detector and dataset we consider. Our locally-calibrated Fast-DetectGPT improves from $0.63$ to $0.85$ AUROC on GPT-5.4 text, while our locally-calibrated DMAP detector achieves state-of-the-art performance. Critically, these gains are achieved with a deliberately naive implementation, leaving substantial further headroom identified in Section~\ref{sec:tuning}. 

The central contribution of this paper is not a new detector. It is a precise diagnosis of a way that existing detectors underperform, a Bayes-optimal framework for correcting this, and a demonstration that even the simplest instantiation of that framework yields dramatic gains. As frontier models continue to close the gap with human writing, we hope this work provides the community with both the conceptual tools and a practical foundation to meet the challenge.

\section{Limitations and Societal Impact}\label{sec:limitations}

Our method is supervised and its calibration is generator- and domain-dependent. Generator-specific training gives the strongest results; in the pooled-generator Modern RAID setting, DALD exceeds calibrated DMAP (AUROC $0.924$ vs.\ $0.904$; Table~\ref{tab:intel_modern_multigenerator_partial}). Under leave-one-domain-out shift, AUROC typically drops by about 8--12 points (Table~\ref{tab:domain_shift_id_ood}). Deployment therefore requires calibration data that covers the target generators, domains, text lengths, and population factors that can affect score distributions, including non-native-speaker bias. A broad training mixture is a partial remedy, not a guarantee; coverage-aware confidence and abstention are important future work. We do not evaluate robustness to adversarial attacks, nor do we evaluate non-English text. The absence of a public longitudinal frontier-model dataset also means that comparisons across Classic and Modern RAID are potentially confounded by differences in output length and sampling settings.

We believe that, appropriately used, detectors of machine-generated text are a net positive for society. That said, we emphasise that care must be taken in deployment, particularly with regard to ensuring that different demographics are not unfairly penalised and in understanding the possibility of false positives. Therefore, in practice, the continual monitoring of any detection system is a critical safeguard, and detector outputs should not be used as sole evidence of authorship.

\begin{ack}

\paragraph{Funding in direct support of this work.}
The first author completed this work while a consultant at the Risk and Security AI Lab, Visa Inc., and was funded through a Visa Inc.\ AI Fellowship scheme. The remaining authors are full-time employees of Visa Inc. All other resources supporting this work, including computing hardware (GPUs), were provided by Visa Inc.

\paragraph{Competing interests.}
The first author completed this work while on unpaid leave from the University of Manchester, which did not fund or contribute to this work. He completed this work while a consultant at Visa Inc.\ and is now employed by QuTwo, which had no role in this work. The remaining authors are employees of Visa Inc., their primary affiliation. The authors declare no other financial relationships or competing interests.
\end{ack}

\bibliographystyle{unsrt}
\bibliography{main}


\appendix

\section{Why One Should Expect Naive Token-Score Aggregation to Underperform}\label{sec:ThreePoints} In the introduction we made three claims motivating Simpson's paradox based on previous results in the literature. In this section we expand on those claims, give more detail about how they are supported by the literature, and give evidence from our experiments to support them directly.

\subsection{Instruction-Tuned Models Satisfy the Likelihood Hypothesis Because of Overconfidence} The article \cite{kemptondmap} gives a method of viewing how a text sits in the next-token probability distributions of a language model. DMAP plots allow one to see how different parts of the detector model's next-token probability distribution are over- or under-represented in the text, compared to the predictions of the detector model. Figures 2 and 8 of \cite{kemptondmap} show that instruction-tuned models such as GPT-3.5 or Mistral-7B Instruct systematically over-sample tokens which the detector model finds likely, leading to a situation where likelihood-based detectors of machine-generated text are effective. By contrast, pretrained language models such as Mistral-7B oversample tokens which the detector model finds unlikely, leading to a situation where likelihood-based detectors of machine-generated text perform worse than a coin-toss. See in particular Table 1 of \cite{kemptondmap}, where it was shown that DetectGPT, Fast-DetectGPT and Binoculars all have AUROC well below 0.5 for detecting texts made by pure sampling from Llama-3.1-8B, Mistral-7B or Qwen3-8B. This was linked to the phenomenon of overconfidence in instruction-tuned models.

\subsection{Overconfidence Varies Across Hidden Space} In the introduction we justified the claim that the phenomenon of token-level overconfidence is not uniform across hidden space by appealing to the work \cite{xie2024calibrating}, where it was shown that temperature-based recalibration of instruction-tuned models works best when the temperature used for recalibrating each token generation step varies with the position in hidden space.

Here we provide direct evidence. We use the training set from Modern RAID with human and GPT-5.4 completions and use $k$-means clustering to divide our PCA projection of hidden space into fifty regions. We then plot over the test set the average value of log-likelihood of GPT-5.4-generated tokens and human-generated tokens for each region of hidden space. Results are shown in Figure~\ref{fig:cluster_heatmap}. 

We see that, while it is generally true that GPT-5.4-generated tokens have higher log-likelihood than human-written tokens, the values vary dramatically between different regions of hidden space, and some regions even have GPT-5.4 tokens having lower log-likelihood than human-written tokens.

\subsection{Human and Machine-Generated Text Occupy Different Parts of Hidden Space}
We previously justified the idea that machine and human text occupy different regions of hidden space by noting the work \cite{chen2025repreguard} which used this idea to build a detector of machine-generated text. To confirm the hypothesis, we analyse the distributions of human and machine-generated contexts among the fifty clusters of hidden space generated for our previous experiment. We show the proportions of GPT-5.4 and human-generated text in each cluster in Table~\ref{tab:top_clusters}.

\begin{table}[htbp]
\centering
\caption{Hidden-space cluster statistics on Modern RAID (human vs.\ GPT-5.4). Hidden vectors are partitioned into 50 regions by $k$-means; the 10 largest clusters are shown. There are clear differences in both the proportion of each cluster corresponding to human and GPT-5.4 tokens, and the mean values of the log-likelihood of human and GPT-5.4 generated tokens within each cluster. Differences between clusters are stronger than differences between GPT-5.4 and human text.}
\label{tab:top_clusters}
\vspace{0.5em}
\resizebox{\textwidth}{!}{
\begin{tabular}{ccrrrr}
\toprule
Cluster ID & Tokens in Cluster & Prop. Human & Prop. GPT-5.4 & Mean $\log p$ (Human) & Mean $\log p$ (GPT-5.4) \\
\midrule
4 & 120,057 & 0.511 & 0.489 & -1.778 & -1.182 \\
21 & 76,984 & 0.514 & 0.486 & -1.555 & -1.386 \\
2 & 57,594 & 0.557 & 0.443 & -3.080 & -2.756 \\
5 & 54,897 & 0.362 & 0.638 & -3.083 & -3.306 \\
11 & 46,805 & 0.440 & 0.560 & -3.353 & -2.877 \\
49 & 44,854 & 0.449 & 0.551 & -2.417 & -1.857 \\
15 & 44,465 & 0.135 & 0.865 & -3.792 & -3.874 \\
13 & 43,404 & 0.443 & 0.557 & -4.129 & -4.652 \\
38 & 43,183 & 0.286 & 0.714 & -2.270 & -2.132 \\
31 & 42,547 & 0.445 & 0.555 & -4.112 & -3.324 \\
\bottomrule
\end{tabular}}
\end{table}

\subsection{{Pair-Level Simpson-Like Reversals}}\label{app:pairwise_reversals}
{Throughout this document, we have used ``Simpson's paradox'' as shorthand both for strict Simpson's paradoxes and for cases in which a strong local signal is significantly weakened by aggregating heterogeneous conditional score distributions. Nevertheless, at the level of pairs of texts, we do see cases of genuine Simpson's paradox.}

{For each matched human/AI pair, we orient the detector scores so that larger values indicate machine text. An ordering is correct when the AI text scores higher than its paired human text. We then compare the uncalibrated and calibrated orderings. ``Both correct'' and ``Both incorrect'' mean that calibration leaves the pair's ordering status unchanged. ``Calibration fixes error'' is a forward reversal, in which the uncalibrated ordering is wrong but the calibrated ordering is correct; ``Calibration introduces error'' is the converse. Table~\ref{tab:pairwise_reversals} reports these four outcomes for each detector and generator using the first 200 tokens of Modern RAID.}

\begin{table}[ht!]
\centering
\caption{{Paired ordering outcomes for calibrated and uncalibrated detectors on the first 200 tokens of Modern RAID. A wrong ordering occurs when the human text scores higher than its paired AI text. Percentages are rounded to one decimal place; the final row averages across all 15 method--generator combinations.}}
\label{tab:pairwise_reversals}
\vspace{0.5em}
{
\small
\begin{tabular}{llrrrr}
\toprule
Method & Generator & \shortstack{Both\\correct} & \shortstack{Calibration\\fixes error} & \shortstack{Calibration\\introduces error} & \shortstack{Both\\incorrect} \\
\midrule
Log-Surprisal & GPT-5.4  & 59.1\% & 30.1\% & 3.2\%  & 7.6\% \\
Log-Surprisal & Gemini & 47.3\% & 35.1\% & 6.1\%  & 11.5\% \\
Log-Surprisal & Claude & 67.7\% & 20.9\% & 5.4\%  & 6.0\% \\
\midrule
Log-Rank & GPT-5.4  & 54.7\% & 33.5\% & 4.1\%  & 7.8\% \\
Log-Rank & Gemini & 41.4\% & 42.8\% & 5.2\%  & 10.6\% \\
Log-Rank & Claude & 64.5\% & 25.2\% & 4.9\%  & 5.4\% \\
\midrule
Token Fast-DetectGPT & GPT-5.4  & 59.7\% & 27.6\% & 5.7\%  & 7.1\% \\
Token Fast-DetectGPT & Gemini & 65.8\% & 15.9\% & 10.0\% & 8.3\% \\
Token Fast-DetectGPT & Claude & 76.5\% & 11.6\% & 7.4\%  & 4.5\% \\
\midrule
Token NPR & GPT-5.4  & 56.4\% & 30.0\% & 6.4\%  & 7.3\% \\
Token NPR & Gemini & 55.1\% & 20.6\% & 15.7\% & 8.6\% \\
Token NPR & Claude & 73.8\% & 15.3\% & 7.2\%  & 3.7\% \\
\midrule
DMAP & GPT-5.4  & 70.8\% & 25.8\% & 1.8\% & 1.7\% \\
DMAP & Gemini & 76.7\% & 19.1\% & 2.9\% & 1.2\% \\
DMAP & Claude & 81.7\% & 16.0\% & 1.3\% & 1.0\% \\
\midrule
\textbf{Average} & \textbf{All} & \textbf{64.3\%} & \textbf{24.5\%} & \textbf{5.5\%} & \textbf{5.8\%} \\
\bottomrule
\end{tabular}
}
\end{table}

{Calibration fixes errors more often than it introduces them for every method--generator combination. Averaged across the 15 combinations, calibration corrects the ordering of 24.5\% of pairs and disrupts the ordering of 5.5\%, giving a net improvement of 19.0 percentage points. These reversals show directly that conditioning token scores on hidden-space features before aggregation can recover pairwise ordering information lost by raw aggregation. We do not observe an analogous reversal at the whole-dataset level; the broader argument of this paper requires only heterogeneous conditional score distributions and the resulting loss of signal under aggregation.}
\section{Scoring Functions}\label{sec:scoring}
Our experiments involve calibrated versions of five detection methods: DMAP, log-likelihood, log-rank, Fast-DetectGPT, and NPR. The first three of these are true token-level scorers, which score a text by aggregating scores of tokens. DMAP is described in Section~\ref{sec:AdaptingDMAP}. Log-likelihood computes $\frac{1}{n}\sum_{i=1}^n \log( p(w_i|w_1\cdots w_{i-1}))$ where $p$ is the model probability from our detector model. Log-rank instead aggregates $\log r(w_i|w_1\cdots w_{i-1})$ where this function gives the position of $w_i$ in the list of possible choices of next token ordered by decreasing model probability $p(\cdot|w_1\cdots w_{i-1})$.  Fast-DetectGPT and NPR are discussed below.
\subsection{Fast-DetectGPT}
Fast-DetectGPT and NPR are not aggregates of token-level scores, and so we need to adapt them in order to fit them into our pipeline. Fast-DetectGPT works by first computing a token score \[g(w_i|w_1\cdots w_{i-1})=\log(p(w_i|w_1\cdots w_{i-1}))-\sum_{v\in V}p(v|w_1\cdots w_{i-1})\log (p(v|w_1\cdots w_{i-1})).\] The second term here is the entropy of the next-token probability distribution, or equivalently, the expected value of the negative log-likelihood of a token randomly chosen according to the detector model. It gives a clever way of contextualising the log-likelihood of the true next token in the text in terms of the log-likelihood of alternatives. However, the final Fast-DetectGPT score is not obtained by aggregating token-level scores; instead, one divides by the standard deviation of the token-level scores. This is problematic; see Section~\ref{sec:text_length_bias}. We recover a token-level approximation to Fast-DetectGPT by computing
\begin{align*}
&\mathrm{FD}_{\text{tok}}(w_1\cdots w_n)\\&:=\frac{1}{n}\sum_{i=1}^n\left(\log(p(w_i|w_1\cdots w_{i-1}))-\sum_{v\in V}p(v|w_1\cdots w_{i-1})\log (p(v|w_1\cdots w_{i-1}))\right). 
\end{align*}
Our token-level approximation to the true Fast-DetectGPT usually underperforms the true version, but after calibration our token-level approximation greatly outperforms the baseline.
\subsection{NPR}
NPR works as follows. Given a text $\underline w =w_1\cdots w_n$ one first generates $k$ perturbations of the original text using the T5 mask-filling model; call these $\underline w^{(j)}$. For the original text and for each of the perturbations compute the per-token log-rank. Finally output
\[
\mathrm{NPR}(\underline w)=\dfrac{\text{per-token log-rank}(\underline w)}{\frac{1}{k}\sum_{j=1}^k \text{ per-token log-rank}(\underline w^{(j)})}.
\]
This is not an aggregate of token-level scores. We approximate the NPR idea as follows. Instead of dividing by expected log-rank of perturbations of the whole text, we subtract the expected log-rank of perturbations of each token, following the method of Fast-DetectGPT. That is, we compute token-level scores
\[
g(w_i|w_1\cdots w_{i-1})= \log(r(w_i|w_1\cdots w_{i-1}))-\sum_{v\in V} p(v|w_1\cdots w_{i-1})\log r(v|w_1\cdots w_{i-1}).
\]
We then compute
\begin{align*}
&\mathrm{NPR}_{\text{tok}}(w_1\cdots w_n)\\
&=\frac{1}{n}\sum_{i=1}^n\left(\log(r(w_i|w_1\cdots w_{i-1}))-\sum_{v\in V} p(v|w_1\cdots w_{i-1})\log r(v|w_1\cdots w_{i-1})\right).
\end{align*}
This is intended to approximate the idea of NPR with a token-level scorer amenable to our calibration methods. As with Fast-DetectGPT, we see that replacing NPR with its token-level approximation does not always increase the score, but after calibration the token-level approximation is much better than the original NPR baseline.

\section{Adapting DMAP to Detect Machine-Generated Text}\label{sec:AdaptingDMAP}

DMAP \cite{kemptondmap} was introduced as a variant of log-likelihood and log-rank with more predictable statistical properties and greater robustness to extremes. We adapt it here to produce a detector of machine-generated text.

Recall that given a text $w_1\cdots w_n$ and a language model $p$, the rank of token $w_i$ is its position in the vocabulary ordered by decreasing probability under $p(\cdot|w_1\cdots w_{i-1})$, with the most probable token having rank 1. This notion of position is blind to the actual probabilities involved. DMAP instead defines $a_i$ as the total probability mass assigned by $p$ to all tokens it considers more likely than $w_i$. The value $a_i$ lies in $[0,1]$ and equals $0$ whenever $w_i$ is the most probable next token. Setting $b_i := a_i + p(w_i|w_1\cdots w_{i-1})$, the interval $[a_i, b_i] \subset [0,1]$ has length equal to the probability of $w_i$ and its position within $[0,1]$ reflects how far down the next-token distribution $w_i$ sits. Formal statistical properties of DMAP are established in \cite{kemptondmap}.

Given a partition of $[0,1]$ into $k$ bins, one can measure what proportion of $[a_i, b_i]$ lies in each bin. DMAP sums these proportions over $i \in \{1, \ldots, n\}$ to produce a histogram.

In our adaptation, we are particularly interested in the extent to which a text contains tokens drawn from the tail of the probability distribution. We partition $[0,1]$ into six bins: $[0, 0.5]$, $[0.5, 0.75]$, $[0.75, 0.9]$, $[0.9, 0.95]$, $[0.95, 0.975]$, and $[0.975, 1]$.  The resulting histogram is normalised by dividing each bin value by the bin width, so that text generated by pure sampling from $p$ produces a flat histogram. The choice of partition can be considered a hyperparameter, and we expect tuning to further increase performance.

\begin{figure}[!t]
    \centering
    \includegraphics[width=0.8\linewidth]{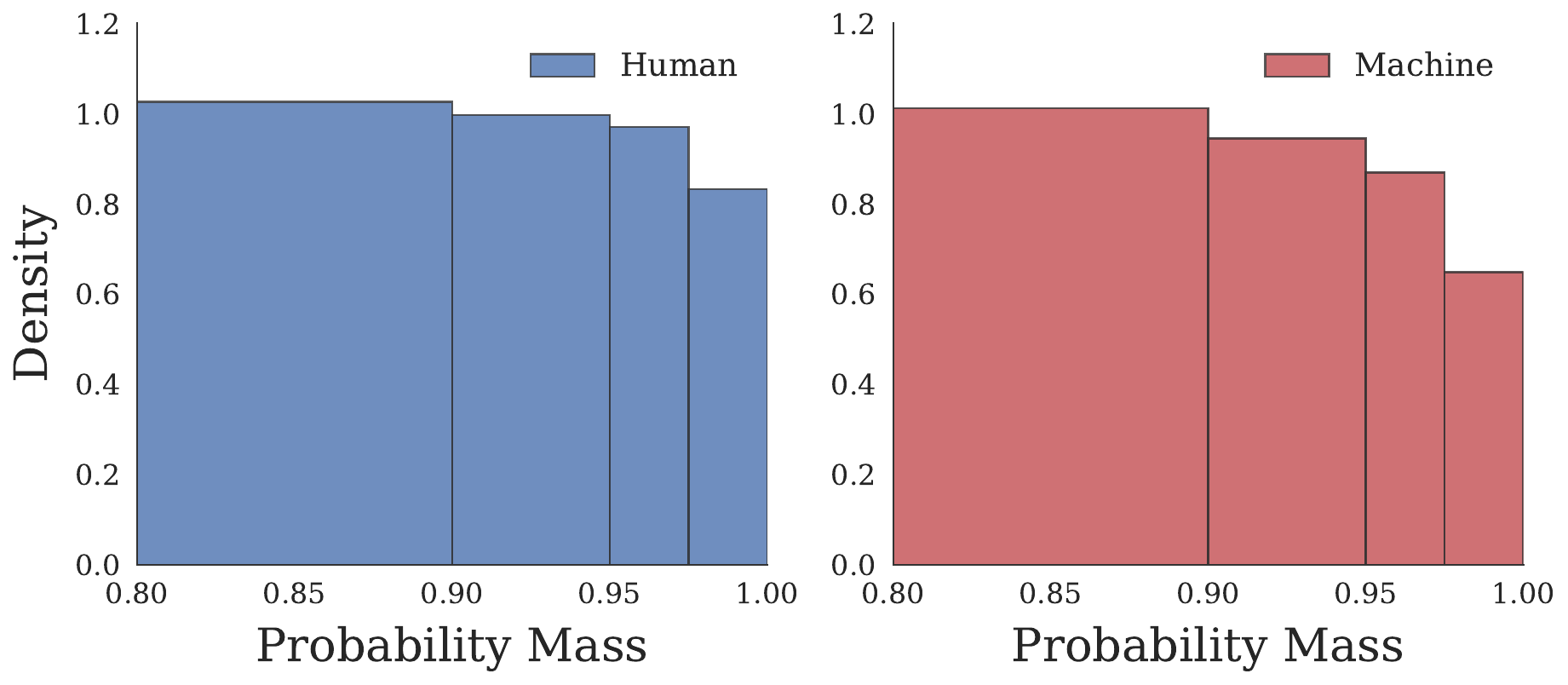}
    \caption{DMAP plots of human-written and GPT-5.4-generated text from the Modern RAID dataset focusing on the tail of the distribution $(0.8,1)$. This highlights the region of biggest difference between human and GPT-5.4 through the lens of the OPT-125M distribution.}\label{fig:global_dmap}
\end{figure}
To score a test text, we first compute DMAP histograms over a training set of human-written and machine-generated texts, yielding representative length-six vectors $q_{\text{human}}$ and $q_{\text{model}}$. For each token $w_i$ in the test text, we compute the proportions $q_i = (q_{i,1}, \ldots, q_{i,6})$ of the interval $[a_i, b_i]$ falling in each bin, and measure the cross-entropy of $q_i$ against $q_{\text{human}}$ and $q_{\text{model}}$ respectively. Subtracting one from the other yields a humanness score for the token, which we then aggregate over the text. This is our Global DMAP detector.

As with other detectors, we see a substantial uplift to performance when we calibrate token-level scores by incorporating information about where a context sits in hidden space, as described in Section~\ref{sec:calibration_implementation}.



\section{Experimental Details}\label{app:experimental_details}
We split datasets into train/test, with all texts corresponding to the same prompt ending up in the same split. For each dataset we take approximately $2{,}000$ texts from each generator (including human-generated text) for training and keep the remainder for test. The same training texts are used to run the PCA and train the local calibrators. 

All learned calibrators are trained for 50 epochs. We used the AdamW optimiser, a learning rate of $10^{-3}$, weight decay of $10^{-4}$, batch size of 4096, and dropout 0.1.

95\% confidence intervals are computed using bootstrap resampling with $10{,}000$ iterations.

\paragraph{Local DMAP predictor.} For local-DMAP defined in Appendix~\ref{sec:AdaptingDMAP}, we train a two-layer MLP with input dimension 30 and hidden dimension 64, with GELU activations and dropout between layers. A final linear layer and softmax yield a predicted distribution over the six bins of the DMAP histogram. The network is trained using soft cross-entropy between predicted and target distributions.

\paragraph{Local Gaussian predictors.} For scalar score functions such as log-likelihood and log-rank, we learn a local Gaussian predictor of $g(x_i | Z(x_i), H)$ and $g(x_i | Z(x_i), M)$ separately. The architecture follows that of the local DMAP predictor, with the final linear layer instead producing a pair $(\mu, \sigma)$. We take $\mu$ directly and apply softplus to $\sigma$ to ensure non-negativity. These networks are trained using negative log-likelihood.

\paragraph{Detector model.} In all experiments we use OPT-125M as the detector language model, which may easily be run on consumer hardware. The exception is the Binoculars baseline, which requires a scorer--observer model pair. For Binoculars we follow the original experimental setup of \cite{hans2024spotting}, but using Falcon-3-7B and Falcon-3-7B-Instruct. For DALD, we use OPT-125M as the base model and fine-tune it using the recipe from \cite{zeng2024dald} exactly, using the same training set for DALD that we use for calibrating other detectors.

\paragraph{Training and evaluation protocol.} For each dataset and generator, we hold out a fixed subset of human-written and machine-generated texts for training local predictors. Specifically, we reserve about $2{,}000$ human-written and about $2{,}000$ machine-generated texts per generator for calibration training, with the remainder used for evaluation. We report AUROC as our primary metric, which does not require a fixed threshold and allows comparison across datasets and generators.

\subsection{Datasets}
\paragraph{Classic RAID.} The RAID dataset \cite{dugan2024raid} contains human-written texts across eight domains together with machine-generated counterparts produced by a variety of models available in 2024. We retain the human texts and those generated by GPT-3.5 and GPT-4, as we are interested in outputs of instruction-tuned models. We use texts generated by pure sampling without repetition penalty or evasion attacks. Each entry includes a title field; models are prompted according to domain, for example in the recipes domain the prompt is: \texttt{Write a recipe for \{title\}}. For this dataset we have $5{,}994$ texts ($1{,}998$ per generator) in the train split and $34{,}119$ texts in the test split.

\paragraph{Modern RAID.} To evaluate on outputs of current frontier models, we regenerate responses following the RAID prompt templates exactly, using GPT-5.4 \cite{openai2026gpt54}, Gemini 3.1 Pro \cite{google2026gemini31pro}, and Claude Sonnet 4.6 \cite{anthropic2026claudesonnet46} as generators. We use default settings for all generators; for Gemini this includes top-$p=0.95$. We have $7{,}980$ texts ($1{,}994$-$1{,}998$ per generator) in the train split and $45{,}396$ in the test split.

\paragraph{Peer-Review benchmark.} Our third dataset is the machine-generated peer-review benchmark of \cite{yu2025your}. Models are presented with papers submitted to ICLR 2021--2022 or NeurIPS 2021--2022, together with conference-specific guidelines and review templates, and asked to produce a review. We consider only texts generated by instruction-tuned models: GPT-4o, Claude 3.5 Sonnet, and Gemini 1.5 Pro. Conference-specific formatting is removed and only the main text of each review is retained. We have $8{,}198$ texts ($2{,}045$-$2{,}052$ per generator) in the train split and $22{,}164$ texts in the test split.

\paragraph{Text length.} Our main results analyse detector performance when given only the first 200 tokens of each text. This controlled-length setting facilitates direct comparison across datasets and generators, and ensures that baseline detector scores are not confounded by text-length effects whose expected value may vary with token count. 

\subsection{Reproducing Figure~\ref{fig:cluster_heatmap}}
The upper plots are 2D t-SNE projections \cite{JMLR:v9:vandermaaten08a} of the OPT-125M embeddings of a single text (Human or GPT-5.4-generated) with t-SNE perplexity 20. To smooth the plot, local averages are computed on a $12\times12$ grid and the surface is generated by cubic interpolation. The text is from the poetry domain of Modern RAID. The lower plots simply visualise a heatmap over $2{,}000$ human and $2{,}000$ GPT-5.4-generated texts from Modern RAID of the log probability across 50 clusters obtained via $k$-means on the hidden space.

\section{Further Results}\label{app:further_results}

\subsection{Confidence Intervals}\label{app:cis}
This section presents companion tables to Table~\ref{tab:calibration_comparison} and Table~\ref{tab:baseline_comparison} containing 95\% confidence intervals for each value obtained via bootstrap resampling from our test set with $10{,}000$ iterations.

\begin{table}[t]
\centering
\caption{Companion to Table \ref{tab:calibration_comparison} with 95\% confidence intervals.}
\label{tab:calibration_comparison_cis}
\vspace{0.5em}
\resizebox{\textwidth}{!}{
\begin{tabular}{lcccccc}
\toprule
\multirow{2}{*}{Method}
& \multicolumn{3}{c}{Modern RAID}
& \multicolumn{3}{c}{Peer-Review} \\
\cmidrule(lr){2-4} \cmidrule(lr){5-7}
& GPT-5.4 & Gemini & Claude
& GPT-4o & Gemini & Claude \\
\midrule

Baseline Log-Surprisal
& (0.554--0.570)
& (0.498--0.513)
& (0.624--0.638)
& (0.625--0.646)
& (0.548--0.570)
& (0.504--0.525) \\

Calibrated Log-Surprisal
& (0.915--0.922)
& (0.889--0.897)
& (0.934--0.940)
& (0.951--0.958)
& (0.966--0.971)
& (0.988--0.991) \\

\midrule

Baseline Log-Rank
& (0.536--0.552)
& (0.461--0.476)
& (0.601--0.616)
& (0.592--0.613)
& (0.516--0.538)
& (0.469--0.490) \\

Calibrated Log-Rank
& (0.851--0.861)
& (0.775--0.787)
& (0.864--0.874)
& (0.818--0.834)
& (0.827--0.842)
& (0.905--0.916) \\

\midrule

Baseline Token NPR
& (0.595--0.610)
& (0.679--0.692)
& (0.762--0.774)
& (0.606--0.626)
& (0.574--0.595)
& (0.527--0.548) \\

Calibrated Token NPR
& (0.829--0.840)
& (0.743--0.755)
& (0.851--0.861)
& (0.801--0.817)
& (0.805--0.821)
& (0.894--0.906) \\

\midrule

Baseline Token Fast-DetectGPT
& (0.620--0.635)
& (0.728--0.741)
& (0.794--0.805)
& (0.630--0.650)
& (0.596--0.617)
& (0.535--0.556) \\

Calibrated Token Fast-DetectGPT
& (0.849--0.859)
& (0.800--0.811)
& (0.863--0.873)
& (0.855--0.869)
& (0.854--0.868)
& (0.927--0.937) \\

\midrule

Baseline DMAP
& (0.682--0.696)
& (0.773--0.785)
& (0.800--0.811)
& (0.826--0.841)
& (0.949--0.956)
& (0.855--0.868) \\

Calibrated DMAP
& (0.936--0.942)
& (0.934--0.940)
& (0.959--0.963)
& (0.971--0.976)
& (0.977--0.981)
& (0.993--0.995) \\
\bottomrule
\end{tabular}
}
\end{table}
\begin{table}[t]
\centering
\caption{Companion to Table \ref{tab:baseline_comparison} with 95\% confidence intervals.}
\label{tab:baseline_comparison_cis}
\vspace{0.5em}
\resizebox{\textwidth}{!}{
\begin{tabular}{lcccccc}
\toprule
\multirow{2}{*}{Method}
& \multicolumn{3}{c}{Modern RAID}
& \multicolumn{3}{c}{Peer-Review} \\
\cmidrule(lr){2-4} \cmidrule(lr){5-7}
& GPT-5.4 & Gemini & Claude
& GPT-4o & Gemini & Claude \\
\midrule

NPR
& (0.519--0.533)
& (0.496--0.510)
& (0.547--0.561)
& (0.591--0.608)
& (0.573--0.591)
& (0.487--0.505) \\

Binoculars
& (0.685--0.699)
& (0.529--0.545)
& (0.621--0.636)
& (0.280--0.298)
& (0.615--0.636)
& (0.497--0.519) \\

Fast-DetectGPT
& (0.622--0.636)
& (0.726--0.739)
& (0.794--0.805)
& (0.630--0.650)
& (0.596--0.617)
& (0.533--0.554) \\

DALD
& (0.890--0.899)
& (0.954--0.959)
& (0.940--0.946)
& (0.845--0.859)
& (0.858--0.871)
& (0.961--0.968) \\

Calibrated DMAP
& (0.936--0.942)
& (0.934--0.940)
& (0.959--0.963)
& (0.971--0.976)
& (0.977--0.981)
& (0.993--0.995) \\

\bottomrule
\end{tabular}
}
\end{table}

\subsection{Classic RAID Results}
We present in Table~\ref{tab:classic_raid_key_metrics} the table of results for the Classic RAID dataset. This contains text generated by older models and is mainly used for performance comparison with more modern models.
\begin{table}[t]
\centering
\caption{{Classic RAID human-versus-AI detection using the first 200 tokens. TPR is reported at 0.1\% and 1\% false-positive rates; entries are point estimates with 95\% bootstrap confidence intervals. Rows are grouped by supervision/calibration regime, and the best value for each generator and metric is bold.}}
\label{tab:classic_raid_key_metrics}
\vspace{0.5em}
\resizebox{\textwidth}{!}{
\begin{tabular}{llccc}
\toprule
Model & Method & TPR@0.1\% & TPR@1\% & AUROC \\
\midrule
\multirow{17}{*}{GPT-3.5} & \multicolumn{4}{l}{{\textit{Zero-shot and uncalibrated score functions}}} \\
& Baseline Log-Surprisal & 3.04 (0.01--9.42) & 18.70 (17.39--20.16) & 0.8519 (0.8470--0.8567) \\
& Baseline Log-Rank & 3.51 (1.20--10.09) & 19.62 (18.21--22.16) & 0.8396 (0.8344--0.8446) \\
& Baseline Fast-DetectGPT & 37.70 (33.04--43.11) & 65.70 (63.61--68.45) & 0.9642 (0.9621--0.9663) \\
& Token Fast-DetectGPT & 32.58 (22.19--39.72) & 69.16 (66.27--71.23) & 0.9666 (0.9645--0.9686) \\
& Token NPR & 25.31 (15.91--37.81) & 64.14 (61.65--66.46) & 0.9602 (0.9579--0.9625) \\
& Baseline NPR & 0.52 (0.31--0.71) & 5.06 (4.28--5.82) & 0.7053 (0.6990--0.7116) \\
& Baseline Binoculars & 20.08 (15.26--25.49) & 43.83 (40.57--46.21) & 0.8933 (0.8892--0.8976) \\
& DMAP & 47.54 (40.90--55.65) & 79.06 (76.92--80.89) & 0.9824 (0.9809--0.9838) \\
& \multicolumn{4}{l}{{\textit{Supervised detector-model adaptation}}} \\
& DALD & \textbf{70.77 (66.27--77.35)} & \textbf{89.20 (87.14--90.74)} & \textbf{0.9917 (0.9905--0.9929)} \\
& \multicolumn{4}{l}{{\textit{Supervised local calibration}}} \\
& Calibrated Log-Surprisal & 47.31 (37.96--59.78) & 79.96 (78.17--82.03) & 0.9852 (0.9839--0.9865) \\
& Calibrated Log-Rank & 0.99 (0.05--24.52) & 62.17 (60.27--64.41) & 0.9412 (0.9379--0.9445) \\
& Calibrated Token Fast-DetectGPT & 23.73 (3.24--33.68) & 63.94 (61.99--66.25) & 0.9491 (0.9460--0.9519) \\
& Calibrated Token NPR & 0.18 (0.05--0.26) & 61.07 (56.55--63.73) & 0.9243 (0.9206--0.9279) \\
& Calibrated DMAP & 67.62 (65.16--74.43) & 88.66 (87.60--90.36) & 0.9910 (0.9900--0.9919) \\
\midrule
\multirow{17}{*}{GPT-4} & \multicolumn{4}{l}{{\textit{Zero-shot and uncalibrated score functions}}} \\
& Baseline Log-Surprisal & 1.18 (0.00--5.21) & 10.91 (10.14--11.71) & 0.6342 (0.6269--0.6413) \\
& Baseline Log-Rank & 1.41 (0.37--5.73) & 10.65 (9.93--11.70) & 0.6151 (0.6076--0.6223) \\
& Baseline Fast-DetectGPT & 4.90 (3.70--6.47) & 16.04 (14.66--17.91) & 0.7859 (0.7801--0.7917) \\
& Token Fast-DetectGPT & 2.66 (1.23--4.06) & 15.78 (14.03--17.39) & 0.7867 (0.7809--0.7925) \\
& Token NPR & 1.74 (0.44--4.07) & 15.02 (13.49--16.52) & 0.7696 (0.7635--0.7756) \\
& Baseline NPR & 0.16 (0.07--0.25) & 1.91 (1.55--2.27) & 0.5767 (0.5699--0.5835) \\
& Baseline Binoculars & 1.13 (0.55--1.64) & 5.57 (4.56--6.51) & 0.6319 (0.6247--0.6392) \\
& DMAP & 1.91 (1.17--3.10) & 13.29 (11.89--14.95) & 0.8238 (0.8183--0.8291) \\
& \multicolumn{4}{l}{{\textit{Supervised detector-model adaptation}}} \\
& DALD & 20.62 (17.43--25.69) & 41.89 (38.71--44.26) & 0.9372 (0.9338--0.9405) \\
& \multicolumn{4}{l}{{\textit{Supervised local calibration}}} \\
& Calibrated Log-Surprisal & 13.40 (8.55--15.36) & 32.93 (29.71--35.48) & 0.9023 (0.8985--0.9062) \\
& Calibrated Log-Rank & 0.20 (0.09--1.05) & 20.24 (18.30--22.12) & 0.8145 (0.8091--0.8200) \\
& Calibrated Token Fast-DetectGPT & 0.82 (0.49--6.20) & 21.56 (18.83--23.32) & 0.8290 (0.8238--0.8342) \\
& Calibrated Token NPR & 0.18 (0.07--0.33) & 13.76 (11.69--16.28) & 0.7905 (0.7847--0.7963) \\
& Calibrated DMAP & \textbf{30.79 (22.90--34.72)} & \textbf{51.55 (48.80--54.68)} & \textbf{0.9414 (0.9385--0.9443)} \\
\bottomrule
\end{tabular}
}
\end{table}

\subsection{Full Results for Modern RAID and Peer-Review}
We include in Tables \ref{tab:modern_raid_key_metrics} and \ref{tab:peer_review_key_metrics} results summarised in the main text.

\begin{table}[t]
\centering
\caption{{Modern RAID human-versus-AI detection using the first 200 tokens. TPR is reported at 0.1\% and 1\% false-positive rates; entries are point estimates with 95\% bootstrap confidence intervals. Rows are grouped by supervision/calibration regime, and the best value for each generator and metric is bold.}}
\label{tab:modern_raid_key_metrics}
\vspace{0.5em}
\resizebox{\textwidth}{!}{
\begin{tabular}{llccc}
\toprule
Model & Method & TPR@0.1\% & TPR@1\% & AUROC \\
\midrule
\multirow{17}{*}{GPT-5.4} & \multicolumn{4}{l}{{\textit{Zero-shot and uncalibrated score functions}}} \\
& Baseline Log-Surprisal & 0.74 (0.01--1.73) & 5.65 (4.93--6.43) & 0.5617 (0.5541--0.5695) \\
& Baseline Log-Rank & 0.55 (0.09--1.48) & 4.93 (4.17--6.19) & 0.5440 (0.5364--0.5519) \\
& Baseline Fast-DetectGPT & 1.30 (0.90--2.08) & 8.40 (7.18--9.86) & 0.6290 (0.6217--0.6364) \\
& Token Fast-DetectGPT & 0.66 (0.16--1.13) & 7.86 (6.27--8.98) & 0.6275 (0.6202--0.6350) \\
& Token NPR & 0.90 (0.20--1.33) & 7.87 (6.88--8.96) & 0.6025 (0.5950--0.6101) \\
& Baseline NPR & 0.10 (0.03--0.19) & 1.67 (1.35--2.04) & 0.5259 (0.5190--0.5328) \\
& Baseline Binoculars & 1.75 (1.30--2.83) & 10.53 (9.29--11.59) & 0.6920 (0.6853--0.6989) \\
& DMAP & 0.85 (0.48--1.52) & 6.75 (5.55--7.55) & 0.6891 (0.6821--0.6960) \\
& \multicolumn{4}{l}{{\textit{Supervised detector-model adaptation}}} \\
& DALD & \textbf{30.22 (27.71--33.67)} & 42.70 (40.65--45.50) & 0.8943 (0.8898--0.8988) \\
& \multicolumn{4}{l}{{\textit{Supervised local calibration}}} \\
& Calibrated Log-Surprisal & 7.33 (4.81--9.23) & 32.95 (29.07--36.30) & 0.9182 (0.9146--0.9218) \\
& Calibrated Log-Rank & 2.73 (0.86--6.95) & 25.92 (23.11--28.09) & 0.8559 (0.8508--0.8608) \\
& Calibrated Token Fast-DetectGPT & 3.74 (2.17--5.20) & 19.31 (16.08--22.06) & 0.8538 (0.8488--0.8589) \\
& Calibrated Token NPR & 2.02 (0.34--3.92) & 16.58 (14.89--18.52) & 0.8342 (0.8288--0.8395) \\
& Calibrated DMAP & 29.14 (19.76--34.94) & \textbf{54.95 (52.83--57.48)} & \textbf{0.9388 (0.9358--0.9417)} \\
\midrule
\multirow{17}{*}{Gemini} & \multicolumn{4}{l}{{\textit{Zero-shot and uncalibrated score functions}}} \\
& Baseline Log-Surprisal & 0.02 (0.00--0.04) & 0.11 (0.05--0.19) & 0.5057 (0.4982--0.5132) \\
& Baseline Log-Rank & 0.02 (0.00--0.04) & 0.06 (0.02--0.12) & 0.4683 (0.4609--0.4759) \\
& Baseline Fast-DetectGPT & 0.19 (0.10--0.40) & 2.85 (2.28--3.72) & 0.7329 (0.7263--0.7393) \\
& Token Fast-DetectGPT & 0.15 (0.04--0.31) & 3.10 (2.35--3.83) & 0.7344 (0.7279--0.7408) \\
& Token NPR & 0.36 (0.05--0.58) & 3.52 (2.88--4.23) & 0.6856 (0.6787--0.6924) \\
& Baseline NPR & 0.07 (0.02--0.14) & 0.91 (0.72--1.18) & 0.5031 (0.4961--0.5100) \\
& Baseline Binoculars & 0.38 (0.22--0.72) & 3.32 (2.80--3.84) & 0.5369 (0.5294--0.5445) \\
& DMAP & 0.07 (0.02--0.27) & 1.59 (1.22--2.17) & 0.7791 (0.7731--0.7851) \\
& \multicolumn{4}{l}{{\textit{Supervised detector-model adaptation}}} \\
& DALD & \textbf{34.74 (29.39--38.76)} & \textbf{56.44 (53.58--59.52)} & \textbf{0.9568 (0.9541--0.9593)} \\
& \multicolumn{4}{l}{{\textit{Supervised local calibration}}} \\
& Calibrated Log-Surprisal & 14.03 (10.92--18.46) & 34.73 (32.84--37.89) & 0.8927 (0.8886--0.8968) \\
& Calibrated Log-Rank & 0.56 (0.14--1.69) & 11.58 (10.58--12.80) & 0.7809 (0.7750--0.7868) \\
& Calibrated Token Fast-DetectGPT & 4.33 (2.37--7.27) & 15.60 (14.47--17.57) & 0.8052 (0.7995--0.8108) \\
& Calibrated Token NPR & 0.22 (0.10--0.94) & 10.55 (9.44--11.64) & 0.7488 (0.7425--0.7551) \\
& Calibrated DMAP & 23.13 (20.54--35.51) & 52.21 (49.56--54.71) & 0.9368 (0.9338--0.9398) \\
\midrule
\multirow{17}{*}{Claude} & \multicolumn{4}{l}{{\textit{Zero-shot and uncalibrated score functions}}} \\
& Baseline Log-Surprisal & 0.06 (0.00--0.20) & 2.17 (1.71--2.83) & 0.6309 (0.6237--0.6381) \\
& Baseline Log-Rank & 0.04 (0.00--0.15) & 1.63 (1.23--2.49) & 0.6083 (0.6010--0.6156) \\
& Baseline Fast-DetectGPT & 2.21 (1.53--3.63) & 13.44 (11.81--15.58) & 0.7993 (0.7937--0.8050) \\
& Token Fast-DetectGPT & 1.43 (0.49--2.27) & 13.27 (10.97--14.59) & 0.7996 (0.7939--0.8052) \\
& Token NPR & 1.90 (0.52--2.65) & 12.49 (11.21--13.78) & 0.7680 (0.7621--0.7741) \\
& Baseline NPR & 0.21 (0.07--0.33) & 1.80 (1.45--2.16) & 0.5541 (0.5472--0.5609) \\
& Baseline Binoculars & 1.58 (1.12--2.61) & 9.06 (7.85--9.95) & 0.6283 (0.6213--0.6357) \\
& DMAP & 0.84 (0.42--1.83) & 7.35 (6.35--9.12) & 0.8052 (0.7996--0.8107) \\
& \multicolumn{4}{l}{{\textit{Supervised detector-model adaptation}}} \\
& DALD & 26.97 (20.53--31.32) & 48.49 (45.62--50.90) & 0.9428 (0.9397--0.9458) \\
& \multicolumn{4}{l}{{\textit{Supervised local calibration}}} \\
& Calibrated Log-Surprisal & 8.83 (2.67--18.67) & 49.47 (45.95--52.20) & 0.9369 (0.9338--0.9399) \\
& Calibrated Log-Rank & 0.04 (0.01--4.44) & 29.74 (26.47--33.07) & 0.8691 (0.8643--0.8740) \\
& Calibrated Token Fast-DetectGPT & 2.33 (0.03--4.61) & 32.67 (30.05--35.80) & 0.8682 (0.8634--0.8730) \\
& Calibrated Token NPR & 0.05 (0.01--0.43) & 24.50 (21.40--27.92) & 0.8559 (0.8509--0.8610) \\
& Calibrated DMAP & \textbf{37.34 (31.57--41.38)} & \textbf{63.50 (61.60--66.20)} & \textbf{0.9612 (0.9590--0.9634)} \\
\bottomrule
\end{tabular}
}
\end{table}

\begin{table}[t]
\centering
\caption{{Peer-Review human-versus-AI detection using the first 200 tokens. TPR is reported at 0.1\% and 1\% false-positive rates; entries are point estimates with 95\% bootstrap confidence intervals. Rows are grouped by supervision/calibration regime, and the best value for each generator and metric is bold.}}
\label{tab:peer_review_key_metrics}
\vspace{0.5em}
\resizebox{\textwidth}{!}{
\begin{tabular}{llccc}
\toprule
Model & Method & TPR@0.1\% & TPR@1\% & AUROC \\
\midrule
\multirow{17}{*}{GPT-4o} & \multicolumn{4}{l}{{\textit{Zero-shot and uncalibrated score functions}}} \\
& Baseline Log-Surprisal & 0.00 (0.00--0.00) & 0.52 (0.11--0.94) & 0.6354 (0.6251--0.6456) \\
& Baseline Log-Rank & 0.00 (0.00--0.04) & 0.32 (0.11--0.83) & 0.6021 (0.5917--0.6125) \\
& Baseline Fast-DetectGPT & 0.14 (0.04--0.58) & 2.06 (1.57--2.66) & 0.6397 (0.6296--0.6498) \\
& Token Fast-DetectGPT & 0.13 (0.04--0.56) & 1.97 (1.51--2.64) & 0.6403 (0.6303--0.6504) \\
& Token NPR & 0.13 (0.04--0.41) & 1.80 (1.37--2.34) & 0.6157 (0.6056--0.6259) \\
& Baseline NPR & 0.16 (0.01--0.37) & 1.82 (1.18--2.32) & 0.5996 (0.5907--0.6083) \\
& Baseline Binoculars & 0.02 (0.00--0.07) & 0.31 (0.16--0.50) & 0.2887 (0.2795--0.2982) \\
& DMAP & 0.49 (0.18--1.04) & 7.95 (6.28--10.17) & 0.8337 (0.8262--0.8411) \\
& \multicolumn{4}{l}{{\textit{Supervised detector-model adaptation}}} \\
& DALD & 8.51 (3.80--12.93) & 27.42 (24.03--30.55) & 0.8523 (0.8453--0.8593) \\
& \multicolumn{4}{l}{{\textit{Supervised local calibration}}} \\
& Calibrated Log-Surprisal & 32.71 (17.33--43.59) & \textbf{64.35 (60.86--66.76)} & 0.9544 (0.9508--0.9580) \\
& Calibrated Log-Rank & 0.22 (0.02--1.23) & 13.72 (10.78--17.97) & 0.8262 (0.8182--0.8340) \\
& Calibrated Token Fast-DetectGPT & 1.48 (0.27--6.68) & 24.35 (19.87--29.50) & 0.8623 (0.8552--0.8693) \\
& Calibrated Token NPR & 1.21 (0.34--3.56) & 14.84 (12.43--18.04) & 0.8089 (0.8007--0.8172) \\
& Calibrated DMAP & \textbf{33.04 (27.94--42.72)} & 63.43 (57.86--66.33) & \textbf{0.9734 (0.9709--0.9758)} \\
\midrule
\multirow{17}{*}{Gemini} & \multicolumn{4}{l}{{\textit{Zero-shot and uncalibrated score functions}}} \\
& Baseline Log-Surprisal & 0.00 (0.00--0.00) & 0.29 (0.07--0.62) & 0.5590 (0.5483--0.5698) \\
& Baseline Log-Rank & 0.00 (0.00--0.02) & 0.25 (0.09--0.61) & 0.5269 (0.5162--0.5377) \\
& Baseline Fast-DetectGPT & 0.23 (0.07--0.61) & 1.89 (1.35--2.50) & 0.6063 (0.5958--0.6168) \\
& Token Fast-DetectGPT & 0.22 (0.07--0.59) & 1.86 (1.29--2.52) & 0.6067 (0.5962--0.6172) \\
& Token NPR & 0.18 (0.05--0.38) & 1.52 (1.07--2.04) & 0.5843 (0.5737--0.5950) \\
& Baseline NPR & 0.11 (0.00--0.29) & 1.28 (0.79--1.75) & 0.5816 (0.5725--0.5906) \\
& Baseline Binoculars & 0.74 (0.36--1.36) & 4.04 (3.13--5.49) & 0.6255 (0.6152--0.6358) \\
& DMAP & 5.76 (4.45--9.94) & 37.81 (31.73--43.03) & 0.9523 (0.9486--0.9559) \\
& \multicolumn{4}{l}{{\textit{Supervised detector-model adaptation}}} \\
& DALD & 12.08 (6.51--19.41) & 33.75 (30.40--38.79) & 0.8646 (0.8577--0.8712) \\
& \multicolumn{4}{l}{{\textit{Supervised local calibration}}} \\
& Calibrated Log-Surprisal & 38.08 (26.44--41.23) & 67.73 (63.40--70.19) & 0.9686 (0.9658--0.9713) \\
& Calibrated Log-Rank & 0.92 (0.38--2.82) & 18.50 (16.39--23.43) & 0.8346 (0.8271--0.8421) \\
& Calibrated Token Fast-DetectGPT & 2.73 (1.68--7.50) & 24.40 (21.13--29.10) & 0.8609 (0.8540--0.8678) \\
& Calibrated Token NPR & 0.81 (0.31--2.49) & 15.30 (12.13--18.67) & 0.8129 (0.8050--0.8209) \\
& Calibrated DMAP & \textbf{43.96 (34.31--55.98)} & \textbf{71.25 (67.99--74.15)} & \textbf{0.9788 (0.9767--0.9808)} \\
\midrule
\multirow{17}{*}{Claude} & \multicolumn{4}{l}{{\textit{Zero-shot and uncalibrated score functions}}} \\
& Baseline Log-Surprisal & 0.00 (0.00--0.00) & 0.02 (0.00--0.16) & 0.5144 (0.5036--0.5250) \\
& Baseline Log-Rank & 0.00 (0.00--0.00) & 0.05 (0.00--0.25) & 0.4791 (0.4685--0.4897) \\
& Baseline Fast-DetectGPT & 0.04 (0.00--0.18) & 0.56 (0.36--0.87) & 0.5433 (0.5327--0.5538) \\
& Token Fast-DetectGPT & 0.02 (0.00--0.16) & 0.60 (0.36--0.87) & 0.5453 (0.5346--0.5558) \\
& Token NPR & 0.07 (0.00--0.18) & 0.63 (0.40--0.97) & 0.5372 (0.5265--0.5476) \\
& Baseline NPR & 0.11 (0.00--0.33) & 0.96 (0.70--1.35) & 0.4958 (0.4867--0.5051) \\
& Baseline Binoculars & 0.25 (0.11--0.51) & 1.66 (1.20--2.27) & 0.5077 (0.4968--0.5186) \\
& DMAP & 0.87 (0.50--2.29) & 13.52 (10.96--15.70) & 0.8613 (0.8545--0.8680) \\
& \multicolumn{4}{l}{{\textit{Supervised detector-model adaptation}}} \\
& DALD & 38.11 (29.86--46.34) & 68.19 (65.15--70.97) & 0.9646 (0.9613--0.9677) \\
& \multicolumn{4}{l}{{\textit{Supervised local calibration}}} \\
& Calibrated Log-Surprisal & 59.15 (41.64--69.42) & 83.86 (81.62--86.33) & 0.9897 (0.9882--0.9910) \\
& Calibrated Log-Rank & 3.74 (1.60--8.06) & 26.59 (21.95--36.89) & 0.9104 (0.9047--0.9161) \\
& Calibrated Token Fast-DetectGPT & 11.02 (9.45--23.06) & 46.25 (42.11--51.55) & 0.9322 (0.9271--0.9371) \\
& Calibrated Token NPR & 7.48 (2.67--13.33) & 31.01 (25.28--35.36) & 0.8996 (0.8937--0.9055) \\
& Calibrated DMAP & \textbf{64.02 (57.15--71.36)} & \textbf{88.60 (84.46--91.00)} & \textbf{0.9943 (0.9933--0.9951)} \\
\bottomrule
\end{tabular}
}
\end{table}

\subsection{Multi-Generator Detection}\label{subsec:cross_detector}
{Table~\ref{tab:intel_modern_multigenerator_partial} pools text from three generators and asks whether a text is AI- or human-written without assuming a known generator. On Peer-Review, calibrated DMAP has the highest AUROC. On Modern RAID, however, supervised DALD outperforms calibrated DMAP (AUROC $0.9240$ vs.\ $0.9043$), showing that the generator-specific gains do not transfer uniformly to a pooled setting.}
\begin{table}[t!]
\centering
\caption{{Multi-generator AI-versus-human detection. Each dataset pools human text with output from all three generators, so the detector is not told which generator produced a test example. Calibrated DMAP leads on Peer-Review, whereas supervised DALD leads on Modern RAID (AUROC $0.9240$ vs.\ $0.9043$).}}\label{tab:intel_modern_multigenerator_partial}
\vspace{0.5em}
\resizebox{\textwidth}{!}{
\begin{tabular}{lcccccc}
\toprule
\multirow{2}{*}{Method} & \multicolumn{3}{c}{Peer-Review} & \multicolumn{3}{c}{Modern RAID} \\
& TPR@0.1\% & TPR@1\% & AUROC & TPR@0.1\% & TPR@1\% & AUROC \\
\midrule
Binoculars Baseline & 1.96 & 9.08 & 0.5617 & 0.73 & 5.88 & 0.5677 \\
Fast-DetectGPT Baseline & 6.46 & 20.76 & 0.8040 & 6.68 & 16.82 & 0.7418 \\
Calibrated DMAP & 67.36 & 92.56 & 0.9967 & 2.79 & 22.51 & 0.9043 \\
Log-Likelihood Baseline & 0.00 & 2.44 & 0.7891 & 0.01 & 5.54 & 0.5892 \\
Log-Rank Baseline & 0.00 & 2.44 & 0.7643 & 0.02 & 4.76 & 0.5664 \\
DALD & 13.60 & 36.27 & 0.8733 & 28.44 & 45.71 & 0.9240 \\
\bottomrule \\
\end{tabular}
}
\end{table}

\subsection{Domain-Shift Experiments}

\begin{table}[t]
\centering
\caption{In-domain (ID) vs.\ out-of-domain (OOD) detection performance under topic shift on Modern RAID. Experiments follow a leave-one-domain-out protocol over eight domains: models are trained on samples from 7/8 domains. OOD performance is evaluated on test samples from the held-out domain, while ID performance is evaluated on test samples drawn from the same 7/8 training domains. Each entry reports the mean $\pm$ standard deviation over all domain splits.}
\label{tab:domain_shift_id_ood}
\vspace{0.5em}
\resizebox{\textwidth}{!}{
\begin{tabular}{llcccc}
\toprule
Method & Generator
& ID AUROC & OOD AUROC
& ID TPR@1\% & OOD TPR@1\% \\
\midrule
Calibrated DMAP & GPT-5.4
& $0.9459 \pm 0.0097$ & $0.8403 \pm 0.0705$
& $46.83 \pm 8.10$ & $16.72 \pm 16.07$ \\

& Gemini
& $0.9358 \pm 0.0116$ & $0.8504 \pm 0.0655$
& $31.94 \pm 7.69$ & $12.20 \pm 14.26$ \\

& Claude
& $0.9547 \pm 0.0110$ & $0.8719 \pm 0.0883$
& $50.73 \pm 9.26$ & $20.23 \pm 23.69$ \\
\midrule
Calibrated Log-Surprisal & GPT-5.4
& $0.9574 \pm 0.0111$ & $0.8383 \pm 0.1458$
& $50.03 \pm 10.11$ & $27.32 \pm 28.82$ \\

& Gemini
& $0.9072 \pm 0.0217$ & $0.8562 \pm 0.0634$
& $17.60 \pm 9.14$ & $5.58 \pm 3.88$ \\

& Claude
& $0.9359 \pm 0.0173$ & $0.8419 \pm 0.1265$
& $41.43 \pm 10.38$ & $25.28 \pm 38.47$ \\
\midrule
Calibrated Log-Rank & GPT-5.4
& $0.9209 \pm 0.0107$ & $0.8382 \pm 0.0679$
& $31.15 \pm 7.93$ & $22.54 \pm 16.31$ \\

& Gemini
& $0.8624 \pm 0.0185$ & $0.7614 \pm 0.0795$
& $15.60 \pm 6.29$ & $3.43 \pm 2.62$ \\

& Claude
& $0.9060 \pm 0.0120$ & $0.8188 \pm 0.0815$
& $36.75 \pm 7.55$ & $14.45 \pm 19.34$ \\
\midrule
Calibrated Fast-DetectGPT & GPT-5.4
& $0.9038 \pm 0.0106$ & $0.8065 \pm 0.1184$
& $16.81 \pm 11.14$ & $21.70 \pm 21.46$ \\

& Gemini
& $0.8620 \pm 0.0076$ & $0.7634 \pm 0.0963$
& $12.84 \pm 3.70$ & $3.46 \pm 2.61$ \\

& Claude
& $0.9228 \pm 0.0107$ & $0.8081 \pm 0.1053$
& $38.72 \pm 6.87$ & $16.13 \pm 22.90$ \\
\midrule
Calibrated NPR & GPT-5.4
& $0.8931 \pm 0.0093$ & $0.7994 \pm 0.0874$
& $20.88 \pm 6.89$ & $19.23 \pm 17.52$ \\

& Gemini
& $0.8415 \pm 0.0122$ & $0.7385 \pm 0.0868$
& $15.16 \pm 3.86$ & $3.65 \pm 3.99$ \\

& Claude
& $0.9049 \pm 0.0122$ & $0.8259 \pm 0.0761$
& $39.31 \pm 5.83$ & $17.56 \pm 18.88$ \\
\bottomrule
\end{tabular}
}
\end{table}

{Table~\ref{tab:domain_shift_id_ood} summarises a leave-one-domain-out evaluation on Modern RAID. Moving from in-domain to out-of-domain data typically reduces AUROC by roughly 8--12 percentage points, with substantial variation across held-out domains. This is direct evidence that the learned conditional distributions are domain-dependent. Because this table contains only locally calibrated methods, it does not support a comparison with zero-shot or uncalibrated baselines; its warranted conclusion is that relevant domain coverage is necessary for reliable calibration.}

\clearpage
\subsection{{Calibration Ablations and Token-Count Sensitivity}}\label{app:token_budgets}
{The following ablations use Modern RAID at 200 tokens unless stated otherwise. Table~\ref{tab:random_features} tests whether hidden-space features carry local information, while Tables \ref{tab:token_auroc}--\ref{tab:token_tpr_one} show how performance changes from 50 to 400 tokens. All methods in the token-count tables except DALD are calibrated.}
\begin{table}
\centering
\caption{{Shuffled-feature ablation on Modern RAID at 200 tokens. The randomly calibrated variant shuffles the hidden-space features $Z$ before calibration. Shuffling lowers AUROC on every generator; low-FPR TPR can nevertheless rise slightly, so the conclusion is restricted to AUROC.}}
\label{tab:random_features}
\vspace{0.5em}
{\begin{tabular}{llccc}
\toprule
Generator & Method & TPR@0.1\% & TPR@1\% & AUROC \\
\midrule
\multirow{2}{*}{GPT-5.4} & DMAP & 0.89 & 6.60 & \textbf{0.6888} \\
& Randomly Calibrated DMAP & 0.63 & 4.49 & 0.6639 \\
\midrule
\multirow{2}{*}{Gemini} & DMAP & 0.06 & 1.61 & \textbf{0.7788} \\
& Randomly Calibrated DMAP & 0.26 & 3.59 & 0.7004 \\
\midrule
\multirow{2}{*}{Claude} & DMAP & 0.82 & 7.49 & \textbf{0.8050} \\
& Randomly Calibrated DMAP & 2.02 & 10.97 & 0.7661 \\
\bottomrule
\end{tabular}}
\end{table}

\begin{table}
\centering
\caption{{AUROC versus token count on Modern RAID. All methods except DALD are calibrated.}}
\label{tab:token_auroc}
\vspace{0.5em}
{{\begin{tabular}{llrrrrrr}
\toprule Generator & Method & 50 & 100 & 150 & 200 & 300 & 400 \\ \midrule
\multirow{6}{*}{GPT-5.4} & DMAP & .8417 & .9164 & .9534 & .9725 & .9894 & .9928 \\
& Log-surprisal & .6920 & .8200 & .8651 & .8929 & .9502 & .9648 \\
& Log-rank & .7172 & .8071 & .8448 & .8667 & .9173 & .9263 \\
& Token Fast-DetectGPT & .7006 & .7993 & .8348 & .8575 & .9138 & .9322 \\
& Token NPR & .6895 & .7878 & .8246 & .8466 & .9028 & .9205 \\
& DALD & .7857 & .8464 & .8761 & .8943 & .9074 & .8793 \\ \midrule
\multirow{6}{*}{Gemini} & DMAP & .8452 & .9137 & .9449 & .9621 & .9793 & .9835 \\
& Log-surprisal & .6751 & .7549 & .7829 & .8127 & .8819 & .9100 \\
& Log-rank & .6855 & .7682 & .8006 & .8237 & .8775 & .8959 \\
& Token Fast-DetectGPT & .6899 & .7503 & .7773 & .8028 & .8533 & .8769 \\
& Token NPR & .6250 & .6875 & .7134 & .7433 & .8152 & .8390 \\
& DALD & .8347 & .9084 & .9413 & .9568 & .9655 & .9715 \\ \midrule
\multirow{6}{*}{Claude} & DMAP & .8950 & .9469 & .9695 & .9796 & .9903 & .9901 \\
& Log-surprisal & .7294 & .8129 & .8510 & .8774 & .9178 & .9137 \\
& Log-rank & .7628 & .8343 & .8659 & .8832 & .9136 & .9045 \\
& Token Fast-DetectGPT & .7444 & .8149 & .8523 & .8764 & .9173 & .9080 \\
& Token NPR & .7463 & .8163 & .8499 & .8717 & .9095 & .9037 \\
& DALD & .8796 & .9216 & .9375 & .9428 & .9481 & .9319 \\ \bottomrule
\end{tabular}}}
\end{table}

\begin{table}
\centering
\caption{{TPR at 0.1\% FPR versus token count on Modern RAID. All methods except DALD are calibrated.}}
\label{tab:token_tpr_pointone}
\vspace{0.5em}
{{\begin{tabular}{llrrrrrr}
\toprule Generator & Method & 50 & 100 & 150 & 200 & 300 & 400 \\ \midrule
\multirow{6}{*}{GPT-5.4} & DMAP & .2531 & .3354 & .3744 & .5423 & .7334 & .8218 \\
& Log-surprisal & .0060 & .0224 & .0334 & .0237 & .0435 & .0415 \\
& Log-rank & .0110 & .0506 & .0791 & .1027 & .3453 & .3578 \\
& Token Fast-DetectGPT & .0103 & .0127 & .0159 & .0219 & .1516 & .2531 \\
& Token NPR & .0106 & .0273 & .0411 & .0446 & .2495 & .3462 \\
& DALD & .0667 & .1723 & .2428 & .3005 & .4271 & .3568 \\ \midrule
\multirow{6}{*}{Gemini} & DMAP & .2309 & .3079 & .4046 & .4866 & .5751 & .6861 \\
& Log-surprisal & .0018 & .0061 & .0039 & .0014 & .0347 & .0005 \\
& Log-rank & .0001 & .0028 & .0063 & .0656 & .2416 & .1925 \\
& Token Fast-DetectGPT & .0011 & .0157 & .0275 & .0440 & .1481 & .1369 \\
& Token NPR & .0003 & .0001 & .0001 & .0011 & .0118 & .0212 \\
& DALD & .1280 & .2494 & .2932 & .3384 & .4559 & .4474 \\ \midrule
\multirow{6}{*}{Claude} & DMAP & .3899 & .4563 & .5539 & .5897 & .7637 & .7970 \\
& Log-surprisal & .0161 & .0163 & .0279 & .0069 & .0315 & .0000 \\
& Log-rank & .0014 & .0202 & .0653 & .0782 & .3712 & .3679 \\
& Token Fast-DetectGPT & .0058 & .0348 & .0411 & .0802 & .3355 & .2063 \\
& Token NPR & .0017 & .0028 & .0087 & .0063 & .3288 & .3115 \\
& DALD & .1403 & .2037 & .2412 & .2791 & .4043 & .3963 \\ \bottomrule
\end{tabular}}}
\end{table}

\begin{table}
\centering
\caption{{TPR at 1\% FPR versus token count on Modern RAID. All methods except DALD are calibrated.}}
\label{tab:token_tpr_one}
\vspace{0.5em}
{{\begin{tabular}{llrrrrrr}
\toprule Generator & Method & 50 & 100 & 150 & 200 & 300 & 400 \\ \midrule
\multirow{6}{*}{GPT-5.4} & DMAP & .3522 & .4864 & .6182 & .7210 & .8757 & .9200 \\
& Log-surprisal & .0643 & .1597 & .2363 & .3225 & .5250 & .5425 \\
& Log-rank & .0553 & .1464 & .2063 & .2809 & .5212 & .5193 \\
& Token Fast-DetectGPT & .0596 & .0993 & .1250 & .2000 & .4232 & .5178 \\
& Token NPR & .0628 & .1342 & .1812 & .2368 & .4601 & .5236 \\
& DALD & .1869 & .3267 & .3896 & .4210 & .5196 & .4444 \\ \midrule
\multirow{6}{*}{Gemini} & DMAP & .3516 & .5003 & .5936 & .6762 & .8147 & .8362 \\
& Log-surprisal & .0293 & .0597 & .0911 & .1293 & .2652 & .2908 \\
& Log-rank & .0116 & .0656 & .1387 & .2047 & .3859 & .4268 \\
& Token Fast-DetectGPT & .0343 & .0921 & .1061 & .1519 & .3355 & .3929 \\
& Token NPR & .0077 & .0258 & .0500 & .0595 & .1329 & .2278 \\
& DALD & .2640 & .4125 & .4976 & .5645 & .6643 & .6741 \\ \midrule
\multirow{6}{*}{Claude} & DMAP & .5011 & .6194 & .7157 & .7827 & .8863 & .8766 \\
& Log-surprisal & .0797 & .1357 & .1816 & .2367 & .3631 & .3123 \\
& Log-rank & .0358 & .1689 & .2739 & .3458 & .5367 & .4880 \\
& Token Fast-DetectGPT & .0630 & .1799 & .2405 & .3168 & .5534 & .5485 \\
& Token NPR & .0450 & .1773 & .2620 & .3264 & .5653 & .5253 \\
& DALD & .3050 & .3981 & .4559 & .4870 & .5745 & .4947 \\ \bottomrule
\end{tabular}}}
\end{table}

\clearpage

\section{Examining the Gaussian Assumption of our Calibration Step}\label{sec:gaussian_assumption}
Our calibration step for log-surprisal, log-rank, Fast-DetectGPT and NPR is based on learning Gaussian predictors. More precisely, for each score function (such as log-surprisal),  generator (such as GPT-5.4 or human) and feature vector ($30$-dimensional vector containing information on where the context sits in hidden space and what the five largest predicted next-token probabilities are given context) we predict a pair $(\mu,\sigma)$ such that the corresponding Gaussian models the distribution of score functions for texts from the given generator in contexts with the given feature vector. We used this Gaussian assumption in the implementation of our calibration step as part of our philosophy of implementing the simplest experiments possible in order to demonstrate the correctness of our central idea, that adding the calibration step substantially lifts the performance of detectors. In Section~\ref{sec:tuning} we mentioned that this Gaussian assumption was likely suboptimal, and that a more advanced local model of the distributions would likely lead to performance uplift.

\begin{figure}[!t]
    \centering
    \includegraphics[width=\linewidth]{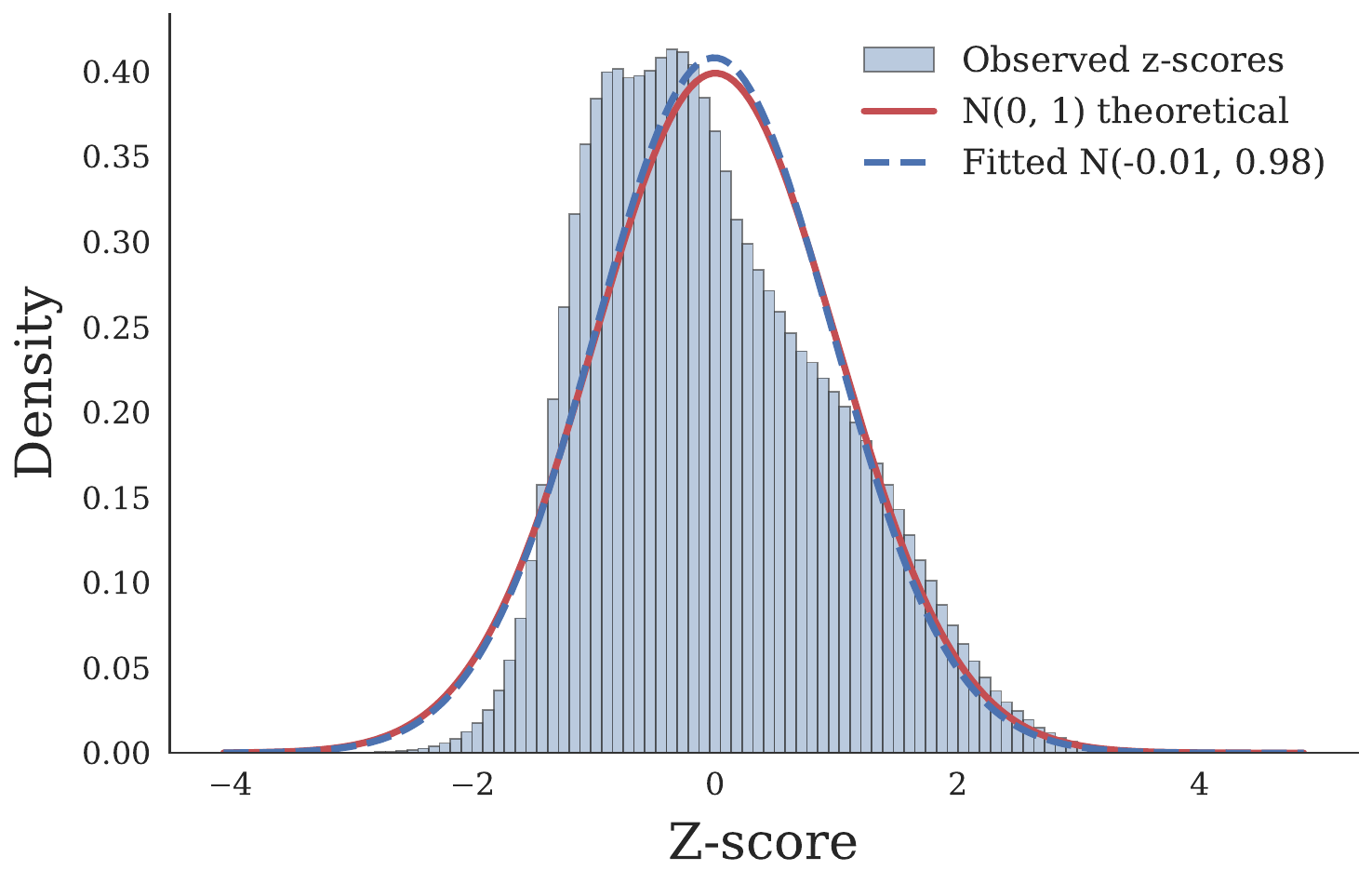}
    \caption{Z-scores of log-surprisal of GPT-5.4-generated text relative to learned calibrating Gaussians. The fact that our observed Z-scores are not well aligned with $\mathcal N(0,1)$ shows a systematic issue with the Gaussian assumption in the local-calibration step; using a more expressive family of distributions for the local calibration is expected to substantially lift performance of our calibrated detectors.}\label{fig:gaussian_miscalibration}
\end{figure}

Here we examine the underperformance of our Gaussian assumption for log-surprisal of GPT-5.4 text. Given a predicted local Gaussian with parameters $(\mu,\sigma)$ and a value $g(w_i)$ of the score of the true next token, we plot the Z-score $\frac{g(w_i)-\mu}{\sigma}$. We run this over all of the texts in the test set, a little over five million tokens in total, and plot our results in Figure~\ref{fig:gaussian_miscalibration}. If the Gaussian distributions matched the underlying data well, the resulting plot of Z-scores would be distributed according to $\mathcal N(0,1)$. We stress that this plot is not assessing the efficacy of individual Gaussian predictors; rather, we aggregate over the test set to see whether there are systematic failures of the local Gaussians to model the underlying data. The clear skew in the histogram, together with the fact that tails are not symmetric, shows that there is clear room for improvement in picking a calibration model.

\section{Degradation of Detector Performance Over Time}\label{app:decay} Our results show clearly that the performance of state-of-the-art detectors at distinguishing human text from that generated by instruction-tuned models is getting worse with more modern models. This is best seen by comparing Classic and Modern RAID (Tables \ref{tab:modern_raid_key_metrics}  and \ref{tab:classic_raid_key_metrics}). Across our baseline detectors designed to extract the likelihood signal, there is degradation in performance as one goes from GPT-3.5 and GPT-4 in the Classic RAID dataset to the models of the Modern RAID dataset. 

\section{Optimising DMAP}\label{app:optimising_dmap}

Table~\ref{tab:optimised_dmap} reports results from a small hyperparameter sweep designed to assess how much headroom remains beyond the minimally tuned configurations used in the main paper. 
The baseline results intentionally reflect a near-default setup and were chosen to isolate the effect of local calibration itself across multiple detectors rather than to optimise absolute performance. 
This section shows that modest optimisation of the calibration step yields stronger performance across all generators. 
We first increase the capacity of the calibration model (raising the hidden dimension of the two-layer MLP from 64 to 256), and then increase the number of DMAP bins to capture finer-grained distributional structure. 
Together, these changes better match the expressivity of the calibration step to the available training data. Finally, we increase the number of training epochs to ensure convergence.

Under both full-context and 200-token evaluation, the optimised DMAP consistently outperforms competing detectors at all reported thresholds, achieving the highest TPR@0.1\% and TPR@1\% for GPT-5.4, Gemini, and Claude on the Modern RAID dataset; see Tables~\ref{tab:optimised_dmap} and~\ref{tab:calibration_supervision}. This demonstrates that the gains reported in the main text are conservative, and that local calibration not only provides a principled correction to likelihood aggregation, but also supports substantially stronger detector variants once modest hyperparameter tuning is applied.

  \begin{table}[t]
  \centering
  \caption{Hyperparameter sweep results for full-context vs.\ 200-token variants.}
  \label{tab:optimised_dmap}
  \vspace{0.5em}
  \resizebox{\textwidth}{!}{
  \begin{tabular}{ll|cccccc}
  \toprule
  \multirow{2}{*}{Setting} & \multirow{2}{*}{Context}
  & \multicolumn{2}{c}{GPT-5.4}
  & \multicolumn{2}{c}{Gemini}
  & \multicolumn{2}{c}{Claude} \\
  & & TPR@1\% & AUROC & TPR@1\% & AUROC & TPR@1\% & AUROC \\
  \midrule
  \multirow{2}{*}{Baseline}
  & Full
  & 68.52\% & 0.9649 & 59.65 & 0.9640 & 70.88 & 0.9763 \\
  & 200 tokens
  & 54.95\% & 0.9388 & 52.21 & 0.9368 & 63.50 & 0.9612 \\
  \midrule
  \multirow{2}{*}{Increase capacity [64 $\rightarrow$ 256]}
  & Full
  & 73.02\% {\small(+4.50)} & 0.9747 {\small(+0.010)}
  & 67.15\% {\small(+7.50)} & 0.9720 {\small(+0.008)}
  & 77.96\% {\small(+7.08)} & 0.9816 {\small(+0.005)} \\
  & 200 tokens
  & 59.35\% {\small(+4.40)} & 0.9513 {\small(+0.013)}
  & 55.40\% {\small(+3.19)} & 0.9461 {\small(+0.009)}
  & 69.11\% {\small(+5.61)} & 0.9679 {\small(+0.007)} \\
  \midrule
  \multirow{2}{*}{Increase num bins [6 $\rightarrow$ 12]}
  & Full
  & 81.19\% {\small(+12.67)} & 0.9828 {\small(+0.018)}
  & 72.37\% {\small(+12.72)} & 0.9742 {\small(+0.010)}
  & 82.14\% {\small(+11.26)} & 0.9857 {\small(+0.009)} \\
  & 200 tokens
  & 67.08\% {\small(+12.13)} & 0.9665 {\small(+0.028)}
  & 62.99\% {\small(+10.78)} & 0.9520 {\small(+0.015)}
  & 73.02\% {\small(+9.52)} & 0.9745 {\small(+0.013)} \\
  \midrule
  \multirow{2}{*}{Increase num epochs [50 $\rightarrow$ 150]}
  & Full
  & 81.10\% {\small(+12.58)} & 0.9840 {\small(+0.019)}
  & 74.53\% {\small(+14.88)} & 0.9769 {\small(+0.013)}
  & 83.61\% {\small(+12.73)} & 0.9871 {\small(+0.011)} \\
  & 200 tokens
  & 70.00\% {\small(+15.05)} & 0.9675 {\small(+0.029)}
  & 65.26\% {\small(+13.05)} & 0.9576 {\small(+0.021)}
  & 74.17\% {\small(+10.67)} & 0.9768 {\small(+0.016)} \\
  \bottomrule
  \end{tabular}
  }
  \end{table}

\section{A Comparison With DALD}\label{app:dald_comparison}

In this section we discuss how our approach differs from DALD \cite{zeng2024dald}. The purpose of DALD is to fine-tune the detector model on generator-model outputs so as to bring the black-box performance of Fast-DetectGPT in line with the white-box performance (where the generator model and detector model are the same). This is well motivated: there is a clear drop off in performance of Fast-DetectGPT between the white-box and black-box cases, and the results of DALD are impressive. 

On the surface, there is some similarity between our method and that of DALD: both are motivated to use outputs of a particular generator model to improve the performance of our detector model. However, the fundamental logic of DALD continues to globally average token-level likelihood signals, specifically, they use Fast-DetectGPT with their fine-tuned detector. This ignores the local structure which our calibrated detectors seek to use. 

One advantage that DALD has is that very little design choice is required: it can use off-the-shelf fine-tuning algorithms and the well-studied Fast-DetectGPT approach. By contrast, our calibration step does require design choices: how long to run for, how large the hidden dimension of the calibration head should be, how many PCA dimensions we should have, and what type of distribution to impose on the local calibration step. Furthermore, the theoretical justification for our method relies on the idea that we have reasonable proxies for the true probabilities of a given token score, given features and generator. 

{Table~\ref{tab:calibration_supervision} provides the direct ablation: supervision without calibration improves DMAP but remains below calibrated DMAP, while combining both is strongest on all three generators. DALD still has the highest Gemini AUROC among the non-combined methods. Appendix~\ref{app:token_budgets} reports the corresponding sensitivity to token budget, which is especially important at low false-positive rates.}

\section{Capping Text Length}\label{sec:text_length_bias}
We note an issue with the way Fast-DetectGPT (and hence DALD) deals with texts of different lengths. This was previously pointed out in \cite{kempton2025temptest}. While the issue is not very easy to see directly from the Fast-DetectGPT paper \cite{baofast}, reading the function \texttt{get\_sampling\_discrepancy\_analytic} from the GitHub repository makes the issue clear. 

While most of our detectors average a token-level score function $g$, Fast-DetectGPT returns $\frac{1}{\sigma}{\sum_{i=1}^n g(x_i)}$. The denominator $\sigma$ is obtained by taking the square root of the sum of token-level variances. This sum of token-level variances grow roughly linearly, or slightly faster once one accounts for covariance. Thus the denominator should grow a little faster than $\sqrt{n}$, while the numerator may be expected to grow linearly in $n$. This introduces a bias when evaluating human vs.\ machine text where the typical length of machine and human text is different. This bias is one of the motivating reasons that all texts are capped to 200 tokens in our main results tables. 

\section{Computation Time} \label{app:compute}

We report the time taken for a complete experimental run on the Peer-Review dataset, which consists of 8,198 training texts (split between human, GPT-4o, Claude, and Gemini) and 22,164 test texts. All experiments requiring GPU acceleration were run on a single NVIDIA A100.

\begin{table}[ht] 
\centering 
\caption{Computation time breakdown for a complete run on the Peer-Review dataset.} 
\label{tab:compute_time} 
\vspace{0.5em}
\begin{tabular}{@{}llr@{}} 
\toprule Stage & Hardware & Time \\ \midrule \multicolumn{3}{l}{\textit{Feature extraction and calibration}} \\ \quad PCA fitting & GPU & 6 min \\ \quad Feature extraction (training set) & GPU & 8 min \\ \quad Feature extraction (test set) & GPU & 24 min \\ \quad Local calibration training (20 runs) & GPU & 70 min \\ \midrule \multicolumn{3}{l}{\textit{Baselines}} \\ \quad Binoculars (200 tokens) & GPU & 3.5 hrs \\ \quad Binoculars (full texts) & GPU & 8.1 hrs \\ \quad NPR (200 tokens) & GPU & 14 hrs \\ \quad Fast-DetectGPT (200 tokens) & GPU & 12 min \\ \midrule \multicolumn{3}{l}{\textit{Other}} \\ \quad Bootstrap CIs \& miscellaneous & CPU & 6 hrs \\ \midrule \textbf{Total (Peer-Review)} & \textbf{GPU / CPU} & \textbf{27h 40m / 6h} \\ \bottomrule \end{tabular} \end{table}

The total runtime is dominated by the NPR and Binoculars baselines, which require expensive forward passes through large language models. Our calibration approach adds minimal overhead: feature extraction and calibration training together require under two hours of GPU time.

Across all three datasets, including preliminary experiments, debugging, and reruns due to data errors, total experimental compute was approximately 220 GPU-hours and 15 CPU-hours.


\end{document}